# Open problems in causal structure learning: A case study of COVID-19 in the UK

Anthony Constantinou[a], Neville K. Kitson[a], Yang Liu[a], Kiattikun Chobtham[a], Arian Hashemzadeh Amirkhizi[b, c], Praharsh A. Nanavati[d], Rendani Mbuvha[a], and Bruno Petrungaro[a].

a. [Bayesian AI](#) research lab, Machine Intelligence and Decision Systems ([MInDS](#)) research group, School of Electronic Engineering and Computer Science (EECS), Queen Mary University of London (QMUL), London, UK, E1 4NS.
b. Sharif University of Technology, Tehran, Iran.
c. University of Toronto, Toronto, Canada.
d. Indian Institute of Science Education and Research, Madhya Pradesh, Bhopal, India.

E-mails: a.constantinou@qmul.ac.uk, n.k.kitson@qmul.ac.uk, yangliu@qmul.ac.uk, k.chobtham@qmul.ac.uk, arian.hashemzadeh@mail.utoronto.ca, praharsh19@iiserb.ac.in, r.mbuvha@qmul.ac.uk, b.petrungaro@qmul.ac.uk.

**Abstract:** Causal machine learning (ML) algorithms recover graphical structures that tell us something about cause-and-effect relationships. The causal representation praovided by these algorithms enables transparency and explainability, which is necessary for decision making in critical real-world problems. Yet, causal ML has had limited impact in practice compared to associational ML. This paper investigates the challenges of causal ML with application to COVID-19 UK pandemic data. We collate data from various public sources and investigate what the various structure learning algorithms learn from these data. We explore the impact of different data formats on algorithms spanning different classes of learning, and assess the results produced by each algorithm, and groups of algorithms, in terms of graphical structure, model dimensionality, sensitivity analysis, confounding variables, predictive and interventional inference. We use these results to highlight open problems in causal structure learning and directions for future research. To facilitate future work, we make all graphs, models, data sets, and source code publicly available online.

## 1 Introduction

*1.1. Causal machine learning*

Research concerned with causal discovery has traditionally focused on experiments, such as Randomised Control Trials (RCTs) that are widely used in clinical research and social sciences. However, it is often expensive and time consuming to perform controlled experiments, if not unethical or impossible. Moreover, while RCTs serve science by estimating the effect of intervention, they do not always produce an accurate estimate of the effect (Deaton and Cartwright, 2018). Therefore, new methods that complement the results obtained from RCTs are often necessary.

Assessing and modelling causal relationships is fundamental to identifying and explaining the causal phenomena we experience. Hypotheses about why certain events happen need to be investigated with measures that go beyond predictive validation, to understand how the world works so that we can become better at influencing it. Nowadays, the necessity of verifiability has led to a revised public understanding of the limitations of black-box Machine Learning (ML) solutions, which tend to provide limited interpretability and explainability. Thus, their ability to support recommended actions for intervention that require explanation is limited (Guidotti et al., 2018; Schölkopf et al., 2021).

Causal ML represents the field of research that broadly focuses on unsupervised learning algorithms whose aim is to recover causal structure from data. These approaches can be used to guide intervention with complete interpretability. While approaches that fall within the field of causal ML have evolved radically over the past few decades, these innovations often come in the form of theoretical advancements with limited impact in practice; at least compared to approaches from associational ML. This is partly because discovering accurate causal relationships represents a notoriously difficult task. This is especially true when working with real data that tend to be imperfect in different ways, as opposed to synthetic experiments. For example, real data may contain missing, biased or incorrect values. It may also be subject to information loss due to dimensionality reduction, have distributions that shift over time, or not capture all of the variables of interest. These imperfections violate many of the assumptions on which many algorithms are built on, and therefore have negative repercussions on the performance of these algorithms (Constantinou et al., 2021).

*1.2. Case-study: COVID-19*

The Coronavirus disease 2019 (COVID-19) pandemic has highlighted the need for determining the effectiveness of unprecedented policy interventions; something which causal models are capable of providing. The first known case of COVID-19 was identified in Wuhan, China, in December 2019. The virus spread worldwide within a few weeks, and the World Health Organisation (WHO) declared COVID-19 a pandemic in March 2020. As of the 11$^{th}$ of October 2022, there have been 23.7M recorded infections and 208K deaths in the UK, and 622M infections and 6.6M deaths worldwide (Our World in Data, 2022a).

To deal with the speed with which COVID-19 was spreading, each country introduced bespoke – although similar - policy interventions that largely involved limiting population mobility and contact. In the UK, the government mandates included social distancing, mask wearing, limiting attendance at various events and other activities, encouraging vaccination, as well as imposing strict lockdowns that would only allow essential mobility. These policies were often informed by national scientific and academic studies (Imperial College COVID-19 Response Team, 2021) that aimed to control hospitalisations.

In an era of big data, there is already an extraordinary number of papers in the academic literature that study virtually all aspects of the COVID-19 pandemic. These studies include research on the relationship between COVID-19 and:

- concentrations of vitamin D, where Hastie et al. (2020) show no evidence that vitamin D explains susceptibility to COVID-19 infection;
- obesity or BMI, where Yates et al. (2020) report early evidence for a dose-response association between BMI, waist circumference and COVID-19;
- the effect of Personal Protective Equipment (PPE), where Nguyen et al. (2020) find that front-line health-care workers are at higher risk of being infected with COVID-19, compared to the general community, and that this is often due to inadequate PPE;
- psychiatric disorders, where Yang et al. (2020) report that individuals with clinically confirmed pre-pandemic psychiatric disorders were at elevated risk of COVID-19 hospitalisation and death;
- clinical, regional, and genetic characteristics, with studies reporting that male sex, lower educational attainment and non-White and Black ethnicities increase the risk of being infected

- with COVID-19 (Chadeau-Hyam et al., 2020) which in turn help predict COVID-19 deaths (Elliot et al., 2021), and Kolin et al. (2020) suggest that discrimination in the labour market may play a role in the high relative risk of COVID-19 for Black individuals and confirm the association of blood type A with COVID-19;
- attributes that predict long COVID, where Sudre et al. (2021) report that early disease features, age and sex largely explain long COVID-19;
- changes in brain structure, where Douaud et al. (2022) report strong evidence of brain-related abnormalities by investigating brain changes, before and after COVID-19 infection, in 785 participants in the UK Biobank.

Other relevant studies include those by O'Connor et al. (2020) who report that UK lockdown appears to have affected the mental health and well-being of the UK adult population, Jarvis et al. (2020) who conclude that physical distancing measures adopted by the UK public reduced contact levels and led to a substantial decline in infections, and Menni et al. (2021) who report that the Pfizer-BioNTech (BNT162b2) and the Oxford-AstraZeneca (ChAdOx1 nCoV-19) COVID-19 vaccines were found to decrease the risk of COVID-19 infection 12 days after vaccination.

While many of the COVID-19 studies report potential causal links between factors of interest, only a few of them employ some sort of causal modelling. These include a) the work by Mastakouri and Schölkopf (2020) who use causal time-series to study the causal relationships amongst German regions in terms of the spread of COVID-19 given the restriction policies applied by the federal states, b) Horn et al. (2020) who studied the causal role of COVID-19 in immunopsychiatry by simulating confounding and mediating regression variables, c) Sahin et al. (2020) who develop a hypothetical causal loop diagram to investigate the complexity of the COVID-19 pandemic and related policy interventions, d) Friston et al. (2020) who construct a dynamic causal model using variational Bayes and hypothetical conditional dependencies applied to epidemiological populations to generate predictions about COVID-19 cases in London, e) Fenton et al. (2020) who explain the need to incorporate causal explanations for the data to avoid biased estimates in COVID-19 testing, and f) Chernozhukov et al. (2021) who construct a hypothetical causal model to investigate the causal impact of masks, policies and behaviour on early COVID-19 pandemic in the U.S. We found only one paper that makes use of causal structure learning; the work by Gencoglu and Gencoglu (2020) on discovering and quantifying causal relationships between pandemic characteristics, Twitter activity, and public sentiment.

### 1.3. Paper purpose and paper structure

In this paper, we focus on COVID-19 data that capture the development of the pandemic in the UK over a period of 2.5 years, with a focus on viral tests, infections, hospitalisations, deaths, vaccinations, policy interventions, and population mobility. These factors are interesting from a causal perspective on the basis that they are assumed to capture causal interactions between them and can serve as a good testbed for causal ML.

An important difference from past COVID-19 studies is that, in this study, we focus on assessing the structure learning algorithms, rather than using the algorithms to learn something about COVID-19. Specifically, the objective is to explore the benefits and limitations of structure learning algorithms with application to a critical real-world case study concerning COVID-19 that can be understood by most readers, and to formulate directions for future research guided by open problems in structure learning.

The paper is structured as follows: Section 2 provides preliminary information on causal models and structure learning, Section 3 describes the process we followed to collect and pre-process COVID-19 data, Section 4 describes the process we followed to obtain a causal graph based on human knowledge, Section 5 describes the structure learning algorithms investigated, Section 6 presents the results, and we provide our concluding remarks in Section 7.

## 2 Preliminaries: causal structure learning and causal modelling

This section provides a brief introduction, without using technical notation, to some of the main concepts of causal modelling and structure learning to help readers understand the approaches used and the findings presented in this paper. For detailed information on structure learning and causal modelling, we direct the readers to the surveys conducted by Koski and Noble (2012) on Bayesian Networks (BNs), by Scanagatta et al (2019) on structure learning, by Kitson et al (2023) who provide a comprehensive review of 74 structure learning algorithms, the book by Koller and Friedman (2009) on probabilistic graphical models, the book by Darwiche (2009) on BNs, and the book by Fenton and Neil (2018) that focuses on knowledge-based BNs.

### 2.1. Causal modelling

A Bayesian Network (BN) is generally represented by a Directed Acyclic Graph (DAG) that contains nodes and directed edges, where each node represents a data variable and each directed edge represents dependency. In discrete BNs, the relationship between variables is captured by Conditional Probability Tables (CPTs), whereas in continuous BNs the relationships are captured by conditional distributions or functional relationships. A Causal Bayesian Network (CBN) is a BN model whose directed edges are assumed to be causal. By extension, this implies that a CBN can only be represented by a single DAG structure, whereas a BN can be represented by any DAG structure that belongs to its corresponding Markov equivalence class of DAG structures, which we cover in subsection 2.2.

A CBN model can be used for predictive and diagnostic inference, but also for interventional and counterfactual inference. This is what Pearl (2018) calls *the ladder of causation*, where the first step focuses on what we can learn from association alone; the second step focuses on the simulation of hypothetical interventions to measure their effect without the need to perform experiments; and the third step on answering counterfactual questions about alternative actions that could have taken place in the past. Pearl's do-operator framework, which represents a mathematical representation and simulation of physical interventions in causal models (Pearl, 2012), is used to model situations that fall within the second step that goes beyond associational inference. In this study, we will explore the first two steps in the ladder of causation.

### 2.2. Causal structure learning

Causal structure learning represents an unsupervised learning process that can be categorised into a) combinatorial optimisation where most algorithms search for a boolean-valued or a discrete-valued adjacency matrix that captures the presence or absence of edges and might include different types of edges, and b) a more recent approach, continuous optimisation, that uses a real-valued adjacency matrix which can be tackled by off-the-shelf optimisers with an acyclicity constraint. Combinatorial optimisation consists of different classes of learning, including:

  i. **Constraint-based** algorithms perform a series of marginal and conditional independence tests, to identify undirected edges between variables and orientate some of those edges.
 ii. **Score-based** algorithms rely on heuristic search techniques to explore the search space of possible graphs, and rely on objective functions to score each graph visited.
     This class of learning can be further subdivided into *approximate* and *exact* solutions. Approximate learning algorithms tend to be efficient but may terminate at a local maxima solution, whereas exact learning algorithms are more computationally expensive in exchange for the guarantee to return the graph that contains the highest score in the search space of graphs. However, this guarantee typically requires that the search space of graphs is limited to a specified node in-degree[1], and graphical structures tend to be explored locally via combinatorial optimisation rather than globally with heuristic search.
iii. **Hybrid** algorithms combine the above two classes of structure learning.

---

[1] The maximum number of parents to be explored per node.

Approaches that rely on continuous optimisation can be categorised into linear and nonlinear score-based solutions. Linear approaches involve constructing linear Structural Causal Models (SCMs) or continuous or functional BNs in which the relationships are linear, and optimise the coefficients in the weight matrix using gradient descent. Nonlinear approaches tend to involve extensions of linear solutions that incorporate neural network models, and allow for the relationships within the adjacency matrix to be nonlinear.

Each class of structure learning algorithms comes with its own set of assumptions, advantages and disadvantages, and there is no consensus in the literature about which learning class might be preferable or suitable. Irrespective of the learning class an algorithm belongs to, different algorithms often produce different types of graphs. This is partly because many algorithms make different assumptions about the input data, and rely on different objective functions. Most of the objective functions considered tend to be score-equivalent; meaning they produce the same score for graphical structures that are part of the same Markov equivalence class. This is because score-equivalent functions assume that not all directed relationships can be recovered from observational data.

Perhaps the most well-known type of graph produced is the Directed Acyclic Graph (DAG), which contains nodes and directed edges that do not allow cyclic relationships. Algorithms that produce a DAG structure use objective functions that are not score-equivalent, or simply return a random DAG of the highest scoring equivalence class. This equivalence class is represented by a Completed Partially Directed Acyclic Graph (CPDAG), which contains undirected (in addition to directed) edges that cannot be orientated from observational data. In other words, a CPDAG represents a set of Markov equivalence DAGs, where each DAG in the equivalence class produces the same objective score.

A more complicated type of graph is the Maximal Ancestral Graph (MAG) that contains bi-directed edges indicating confounding, in addition to directed and undirected edges. Unlike a DAG or a CPDAG, a MAG captures information about possible latent confounders and ancestral relationships[2]. The equivalence class of a MAG is represented by a Partial Ancestral Graph (PAG), which is analogous to the relationship between a DAG and a CPDAG. Algorithms that consider the possibility of latent confounders produce a MAG or a PAG, and tend to fall under the constraint-based or hybrid classes of learning.

### 3. Data collection and data pre-processing

Data were collected from different public sources, with the majority of information coming from official UK government websites for COVID-19. The data set collated contains 18 columns and 866 rows, where each data column corresponds to a variable and each data row to a daily outcome. The data instances capture the progress of the pandemic in the UK over 866 days; from the 30th of January 2020 to the 13th of June 2022.

We describe and categorise the data variables by the type of information they capture in subsection 3.1. The raw data contain both continuous and categorical variables, as well as missing data values. Subsection 3.2 describes how we pre-process the data so that it becomes suitable for input to the different structure learning algorithms considered.

#### 3.1. *Categorisation and description of the data variables*

Table 1 lists all the data variables along with their description and data source. Note that the variable *Date* is added for information only. It cannot be taken into consideration during structure learning since the algorithms assume that the data rows do not have a temporal relationship, and would consider *Date* as a variable that contains as many states as there are samples, which would make no difference to the learnt structure (we discuss this limitation in Section 4 and conclusions). We distribute the data variables into the following eight categories:

---

[2] A MAG may contain other types of edges and can be used to represent *selection* variables. We do not cover these details in Preliminaries since they are out of the scope of this study.

**Viral tests:** A single variable fits this category; the *Tests across all four Pillars*. It captures information about the different types of tests used to test for COVID-19 infection, across testing Pillars 1, 2, 3 and 4 as defined by the UK Government GOV.UK (2022a). Specifically, testing conducted under Pillar 1 represent tests carried out by the National Health Service (NHS) and the UK Health Security Agency (UKHSA), under Pillar 2 by the UK government COVID-19 testing programme, under Pillar 3 involve antibody serology testing, and under Pillar 4 involve testing for national surveillance.

**Infections:** Three variables make up this category; namely *Positive tests*, *New infections*, and *Reinfections*. These variables capture information about daily cases, and whether those cases were new infections or reinfections.

**Hospitalisations:** This category consists of the variables *Hospital admissions*, *Patients in hospital*, and *Patients in MVBs*. These three variables capture information about the number of patients admitted to hospital with COVID-19, the number of patients in hospital with COVID-19, and the number of patients in Mechanical Ventilator Beds (MVBs) with COVID-19.

**Vaccinations:** A single variable, called *Second dose uptake*, that captures information about the 2$^{nd}$ dose vaccine uptake, and which represents the proportion of the eligible population who received the vaccine. In the UK, two doses of vaccine were required for someone to be considered as 'fully vaccinated'. Booster doses were recommended later, but we do not consider them here for simplicity.

**Deaths:** Two variables, called *Deaths with COVID on certificate* and *Excess mortality*, that capture the number of deaths with COVID-19 listed on the death certificate, and overall excess mortality.

**Mobility:** Three variables that capture information about mobility in the UK in the form of indices. These are the *Transportation activity*, *Work and school activity*, and *Leisure activity*. They correspond to combined indices about flights, buses, trains and transit stations, visits to parks, retail, grocery, and restaurants, as well as walking, journeys, homeworking and school activities. As indicated in Table 1, while these data are provided by the UK Government, the data come through third-party providers such as Transport for London (TfL), Google, Apple, Citymapper, and OpenTable. The *Schools* index, that represents one of the indices that make up the *Work and school activity* variable, was constructed manually as described in Table A.4, since we were unable to find this information readily available online.

**Policy:** The two variables, *Face masks* and *Lockdown*, capture important UK policy interventions about lockdowns and mask mandates. Because we were unable to find these data readily available online, we constructed both of these variables manually from the sources specified in Table 1, and as described in Tables A.4 and A.5.

**Other:** The two variables, *Season* and *Majority COVID variant*, which can be viewed as background information that did not fit under any of the previously defined categories. These variables capture information about season (e.g., summer), and the majority COVID-19 variant. They are also constructed manually as described in Table A.4.

arXiv pre-print, 2023.**Table 1.** Description of the raw data variables collated to formulate the COVID-19 UK data set. All of the data variables follow an ordinal distribution of values or states.

| No. | Var name | Category | Var type | Description and data source |
|---|---|---|---|---|
| 0 | Date | n/a | Disc. | The date the observations were recorded. Not considered for structure learning. |
| 1 | Tests across all four Pillars | Viral tests | Cont. | The total number of tests across Pillars 1, 2, 3 and 4 (GOV.UK, 2022a). |
| 2 | Positive tests | Infections | Cont. | Number of cases (i.e., people tested positive for COVID-19) by specimen date (GOV.UK, 2022b). |
| 3 | New infections | Infections | Cont. | New infections by specimen date (GOV.UK, 2022b). |
| 4 | Reinfections | Infections | Cont. | New reinfections by specimen date (GOV.UK, 2022b). |
| 5 | Hospital admissions | Hospitalisation | Cont. | Number of patients admitted to hospital with COVID-19 (GOV.UK, 2022c). |
| 6 | Patients in hospital | Hospitalisation | Cont. | Number of patients in hospital with COVID-19 (GOV.UK, 2022c). |
| 7 | Patients in MVBs | Hospitalisation | Cont. | Number of patients in MVBs with COVID-19 (GOV.UK, 2022c). |
| 8 | Second dose uptake | Vaccines | Cont. | Reported $2^{nd}$ dose vaccination uptake (GOV.UK, 2022d). |
| 9 | Deaths with COVID on certificate | Deaths | Cont. | Daily deaths with COVID-19 on the death certificate by date of death recorded (GOV.UK, 2022e). |
| 10 | Excess mortality | Deaths | Cont. | The percentage difference between the reported number of deaths and the projected number of deaths for the same period based on previous years. (Our World in Data, 2022c) |
| 11 | Transportation activity | Mobility | Cont. | Combined UK government data indices provided by TfL on bus and tube activity, by Google on transit stations activity, by Citymapper on journeys activity, by Apple on walking activity, and UK flight activity (GOV.UK, 2022f; ONS.GOV.UK, 2022; EUROCONTROL, 2022). |
| 12 | Work and school activity | Mobility | Cont. | Combined UK government data indices provided by Google on homeworking and workplace activities, retail and recreation, grocery and pharmacy activities, and schools' operational guidance during COVID-19 in the UK (GOV.UK, 2022f; 2022h; 2022i; 2022j; Wikipedia, 2022b). |
| 13 | Leisure activity | Mobility | Cont. | Combined UK government data indices provided by Google on park visits, and by OpenTable on restaurant bookings (GOV.UK, 2022f). |
| 14 | Face masks | Policy | Categ. | Mask mandates in the UK during the COVID-19 pandemic (GOV.UK, 2022g; Wikipedia, 2022a). |
| 15 | Lockdown | Policy | Categ. | Lockdown mandates in the UK during the COVID-19 pandemic (Institute for Government, 2022) |
| 16 | Season | Other | Categ. | The four seasons; winter, autumn, summer, spring. |
| 17 | Majority COVID variant | Other | Categ. | The majority COVID-19 variant in the UK (GOV.UK, 2022k; Our World in Data, 2022b) |

### 3.2. *Continuous, discrete and mixed, complete and incomplete data sets*

The raw data set contains both continuous and categorical variables, some of which are incomplete. However, many algorithms work with one data format only, and most algorithms assume the input data set is complete without missing data values. To be able to test all algorithms across all possible data inputs they accept, we create the following seven data sets derived from the raw data:

i.  **Discrete quartiles incomplete:** where all continuous variables are discretised into quartiles, and missing values were not imputed.

  ii. **Discrete quartiles complete:** where all continuous variables are discretised into quartiles, and all missing values within the data set are imputed given the discretised values.
  iii. **Discrete k-means incomplete:** where all continuous variables are discretised using k-means clustering, and missing values were not imputed.
  iv. **Discrete k-means complete:** where all continuous variables are discretised using k-means clustering, and all missing values within the data set are imputed given the discretised values.
  v. **Continuous incomplete:** where all categorical variables are converted into continuous variables, and missing values were not imputed.
  vi. **Continuous complete:** where all categorical variables are converted into continuous variables, and all missing values in the data set are imputed given the continuous values.
  vii. **Mixed complete:** The raw data set with missing values imputed.

For example, we did not construct a *Mixed incomplete* data set because no algorithm would accept a mixed data set containing missing values. All seven data sets are made publicly available online and can be downloaded from the Bayesys repository (Constantinou et al., 2020).

The continuous data sets are constructed by converting all categorical variables into a continuous range of values from 0 to 1 that correspond to the ordering of those categories. We use two common approaches to data discretisation. The first involves obtaining the quartile intervals for each continuous variable and converting them into four states $\{Very\ Low, Low, High, Very\ High\}$, where each state corresponds to the appropriate quartile interval. The second approach involves applying unsupervised $k$-means clustering to the values of each variable, to determine these four states based on clustering rather than based on quartiles. We used the *sklearn* version 1.1.2 and set the hyperparameter to $k = 4$ to be consistent with the number of states produced by the former approach, and ensure that the states derived from those clusters are ordered such that the ordering of those states is consistent with the ordering of the states derived from quartiles. These two approaches are contrasting since the quartile-based discretisation leads to balanced distributions whereas clustering does not. That is, a balanced distribution maximises the sample size available to parameterise each parameter in a Conditional Probability Table (CPT), whereas the clustering approach represents classic discretisation where we seek to maximise data fitting.

Lastly, because many missing data values are unlikely to be missing at random, we make use of the Markov Blanket Miss Forest (MBMF) imputation algorithm that performs imputation under the assumptions of both random and systematic missingness. The MBMF algorithm is designed to recover the Markov blanket[3] of partially observed variables using the graphical expression of missingness known as the m-graph, which is a type of graph that captures observed variables together with the potential causes of missingness known as missing indicators. This approach was shown to improve imputation accuracy relative to the state-of-the-art, both under random and systematic missingness (Liu and Constantinou, 2023).

### 4   Constructing a causal graph from knowledge

We start by constructing a causal graph based on human knowledge. The knowledge graph is presented in Figure 3 and contains the 17 variables described in Table 1 (excludes *Date*). It represents a consensus across all the authors. Group discussions were conducted to discuss uncertain relationships and to resolve disagreements. Note that while most authors have prior experience in applying causal models to healthcare, none of the authors is an expert in epidemiology. On this basis, we assume that the knowledge graph presented in Figure 3 represents common knowledge about the COVID-19 pandemic. It was constructed based on the following five key assumptions:

  i. The rate of infection is influenced by both the season and the COVID-19 variant;

---

[3] The Markov blanket of a given variable represents the subset of the variables that can fully explain its values. In a DAG model, the Markov blanket of a given node would contain that node's parents, its children, and the parents of its children.

ii. The risk of hospitalisation is influenced by the COVID-19 variant;
iii. Government policy about lockdowns and face masks, as well as vaccine uptake, are influenced by the rates of COVID-19 hospitalisations and deaths;
iv. Mobility is influenced by government policy on lockdowns, but also by season;
v. Excess mortality is influenced by deaths from COVID-19, but also by lockdowns due to changes in mobility that affected other lives in ways that are difficult to measure.

It is important to highlight that, in constructing the knowledge graph, we faced the problem of circular relationships. For example, the status of the pandemic (e.g., rates of infection and hospitalisation) influences lockdown decisions, but lockdowns and other policy actions are expected to affect the future trajectory of the pandemic. Similarly, the rate of infections influences mobility, but mobility also influences the future rate of infection. We recognise that some of the variables are involved in feedback loops, but since we are not modelling temporal relationships in this study, we are not able to include some of these relationships.

The knowledge graph assumes that the starting point is the pandemic since. For example, we must first observe high rates of hospitalisation before we observe policy intervention (e.g., lockdown). Because the edges drawn in Figure 1 represent same-date relationships, we assume that the status of the pandemic is already influenced by any past policy and mobility not shown in the knowledge graph. For example, if we observe low infections that can be explained by reduced mobility imposed a few weeks prior, we assume that there is no same-date influence from mobility to infections, but that the same-date influence is instead from infections to mobility. This is because the effect of policy intervention, such as lockdown or reduced mobility, is observed later and cannot be found within the same data row; e.g., observing lockdown in the data does not necessarily imply that the same-date infections are influenced by lockdown imposed on that day.

On this basis, we assume that there should not be a same-date edge from policy and mobility to infections and hospitalisations in the knowledge graph. Likewise, we assume that policy is influenced by the current status of the pandemic. Therefore, the same-date relationships depicted in the knowledge graph capture observational events, but not interventional; e.g., intervening on mobility would not enable accurate simulation of lockdown interventions in terms of same-date relationships, since this would require that we link such interventions with future observations.

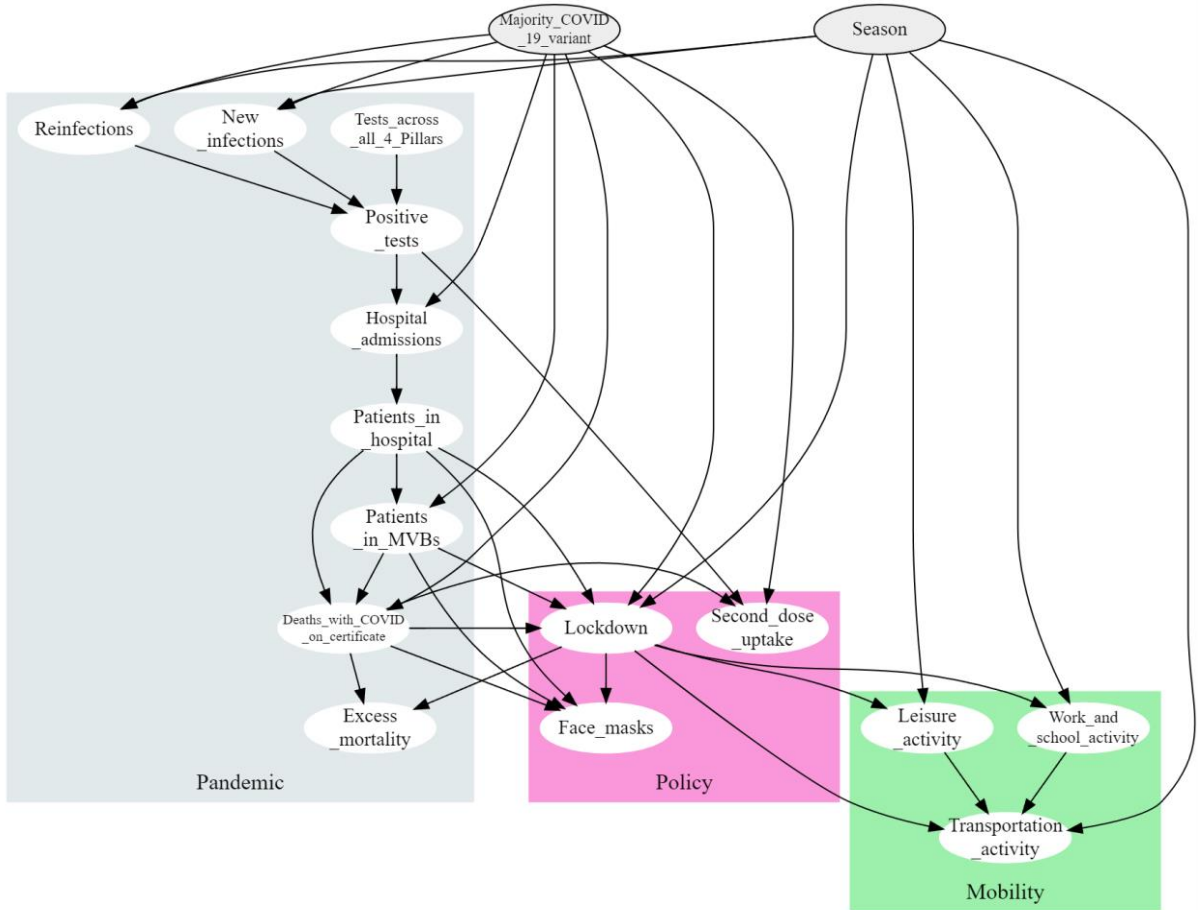

**Figure 1.** The knowledge graph. It contains the 17 variables described in Table 1 (excludes *Date*), 37 directed edges, has a maximum node in-degree of 5, out-degree of 7, and node-degree of 7.

## 5  Learning causal structure from data

We employ structure learning algorithms spanning different classes of learning, that rely on different assumptions about the input data, and which produce different types of graphical structures. Specifically, we consider constraint-based, approximate and exact score-based, hybrid learning, and continuous optimisation algorithms, and which produce either a DAG, CPDAG, MAG or a PAG output (refer to subsection 2.1 for definitions). Moreover, some of these algorithms assume complete or incomplete data, and further assume that the input data are categorical, continuous or mixed (refer to subsection 3.2 for definitions).

Table 2 lists the algorithms along with the assumptions made about the input data. An algorithm is applied to all the different data formats it can process. Moreover, if an algorithm is designed specifically for incomplete data, then it is applied to the incomplete data set rather than to the complete data set. We also consider model-averaging approaches to obtain an overall structure for sets of algorithms that belong to a particular group or structure learning class (details below).

In total, we tested 29 different algorithms, including model-averaging learners, leading to the 64 experiments enumerated in Table 2. The selection of these algorithms ensures that we investigate commonly used algorithms, as well as some less popular algorithms but which come with unique characteristics, across all classes of structure learning. All the algorithms described below return a random DAG or MAG structure from their corresponding equivalence class, if not the actual equivalence class, unless otherwise stated. Detailed technical information about these algorithms is out of the scope of this paper, and can be found in the references provided below for each algorithm, and in the survey papers discussed in Section 2. In alphabetical order, the algorithms we have tested are:

- **CCHM:** A hybrid algorithm by Chobtham and Constantinou (2020) that starts with the approach of CFCI (described below) to learn a skeleton and determine $v$-structures. It then applies hill-climbing to orientate some of the undirected edges in the MAG space, and the do-operator to orientate any remaining undirected edges.
- **FCI:** Fast Causal Inference (FCI) is a constraint-based algorithm that extends PC by accounting for the possibility of latent variables in the data (Spirtes et al., 1999). It starts by producing the skeleton of the graph, followed by conditional independence tests that orientate some of those edges, including producing bi-directed edges indicating confounding, resulting in a PAG output.
- **FGES:** A 'fast' (efficient) and parallelised version of the Greedy Equivalence Search (GES) algorithm by Chickering (2002). The Fast GES (FGES) by Ramsey et al. (2016) contains two learning phases, known as the forward and backward search phases. The forward phase starts from an empty graph and, at each iteration, the edge that maximises the objective function in the CPDAG space is added to the graph. When no edge is found to further increase the objective score, FGES enters the backward phase where edge removals are performed in the same way, and stops when no further edge removals increase the objective score.
- **GFCI:** The Greedy Fast Causal Inference (GFCI) by Ogarrio et al (2016) is a hybrid algorithm that first uses FGES to produce a CPDAG and then performs conditional independence tests to remove extraneous adjacencies in this skeleton, followed by modified FCI orientation rules to produce a PAG.
- **GOBNILP:** An Integer Linear Programming (ILP) score-based algorithm by Cussens (2011) that provides exact learning through a combination of pruning and optimal assignment of parents to each node. ILP guarantees to return the graph that maximises a scoring function, but this guarantee is typically limited to a low maximum node in-degree due to high computational complexity. For example, its default hyperparameters restrict learning to three parents per node to address efficiency issues.
- **HC:** The Hill-Climbing algorithm represents the simplest as well as one of the earliest score-based algorithms for structure learning (Bouckaert, 1994; Heckerman et al., 1995). It starts from an empty graph and investigates all possible neighbouring DAGs by performing arc additions, reversals and removals at each iteration. It then applies the graph modification that maximises the objective score, and stops when the objective score no longer increases.
- **HC_aIPW:** A hybrid algorithm by Liu and Constantinou (2022) that applies the test-wise deletion and Inverse Probability Weights (IPW) to the score-based HC algorithm to more effectively deal with random and systematic missing data in discrete variables.
- **HCLC-V:** The hybrid Hill-Climbing Latent Confounder search with VBEM (HCLC-V) by Chobtham and Constantinou (2022), learns graphs that contain density estimations of possible latent confounders. It employs hill-climbing over the MAG space and uses the p-ELBO (known as the evidence lower bound, or variational lower bound) score as the objective function to determine latent confounders and learn their distributions. The p-ELBO score is not score-equivalent, and so this algorithm returns a unique DAG. HCLC-V returns a DAG structure that may contain latent variables as observed variables. To ensure that the graphs produced by HCLC-V are evaluated in the same way as the other algorithms, we remove any latent variables and introduce the necessary bi-directed edges for latent confounders, thereby obtaining a MAG structure.
- **MAHC:** The Model-Averaging Hill-Climbing algorithm by Constantinou et al. (2022) starts by pruning the search space of graphs to restrict the candidate parents for each node. The pruning strategy represents an aggressive version of those applied to combinatorial optimisation structure learning problems. It then performs model-averaging in the hill-climbing search process and moves to the neighbouring graph that maximises the average objective score, across that neighbouring graph and all its valid neighbouring graphs, under the assumption that a model-averaging process would be less sensitive to data noise and more suitable for real data. Because the model-averaging approach allocates multiple BIC scores to each node and averages them, it results to an objective function that is not score-equivalent and leads to a unique DAG structure.

- **MMHC:** The hybrid Max-Min Hill-Climbing (MMHC) algorithm by Tsamardinos et al (2006) first constructs a skeleton using the constraint-based Max-Min Parents and Children (MMPC) algorithm, and then orientates those edges using HC.
- **NOTEARS:** A continuous optimisation algorithm by Zheng et al. (2018) that converts the traditional score-based combinatorial optimisation problem into an equality-constrained problem with an acyclicity requirement, and returns a unique DAG.
- **NOTEARS-MLP:** An extension of NOTEARS by Zheng et al. (2020) that leverages non-parametric sparsity based on partial derivatives for continuous optimisation, to recover DAGs under the assumption the relationships are nonlinear.
- **Partition-MCMC:** The Partition Markov Chain Monte Carlo (Partition-MCMC) by Kuipers and Moffa (2017) searches the space of partitions, also known as the partial topological ordering of nodes in a DAG. It scores each partition visited by looking at the scores of all DAGs that are consistent with that partition, and returns a random DAG that is consistent with those partitions (not necessarily score-equivalent DAGs).
- **PC-Stable:** The stable version of the classic Peter-Clark (PC) constraint-based algorithm (Spirtes and Glymour, 1991). It starts from a fully connected graph and eliminates edges that lead to marginal or conditional independence. It then orientates some of those edges by performing a set of $v$-structure tests. The PC-Stable variant by Colombo and Maathuis (2014) makes PC insensitive to the order of the variables as they are read from data, by correcting the order of edge deletions. This algorithm generates a Partially DAG (PDAG), which is a simplified version of a CPDAG that also contains directed and undirected edges, with the directed edges indicating the $v$-structures[4].
- **SaiyanH:** A hybrid algorithm by Constantinou (2020) that starts with a denser version of a maximum spanning tree graph, and orientates those edges by performing a sequence of conditional independence tests, BIC maximisation tests, and orientations that maximise the effect of hypothetical interventions. It then performs *tabu* search in the DAG-space with the restriction not to visit graphs that contain disjoint subgraphs. This learning strategy ensures that the learnt model would enable full propagation of evidence.
- **SED:** The Spurious Edge Detection (SED) algorithm by Liu et al. (2022) can be viewed as an additional learning phase that can be applied to the output of any other structure learning algorithm. It takes graphical structures as input and assesses them for possible false-positive edges that could have been produced in the presence of measurement error, and removes them. This is done by looking at 3-vertex cliques that could have been produced in the presence of measurement error. As shown in Table 2, SED was applied to five other score-based, constraint-based and hybrid algorithms.
- **Structural EM:** One of the older algorithms, the Structural EM by Friedman (1997) learns the structure from data that contain missing values. It involves two steps: the Expectation (E) step where missing values are inferred to complete the data, and the Maximisation (M) step where the complete data set is used for structure learning. It returns a random DAG from its equivalence class.
- **TABU:** An extension of HC that performs *tabu* search. It permits the exploration of DAGs that decrease the objective score, and maintains a *tabu* list containing the most recently visited graphs to prevent the algorithm from returning to a graph that was recently visited (Bouckaert, 1995). This approach enables TABU to move into new graphical regions that may contain an improved local maximum compared to HC.

As indicated in Table 2, we used bnlearn v4.8.1 by Scutari (2023) to test HC, TABU, PC-Stable, Structural EM and MMHC, the Tetrad package and its Python extension causal-learn by the Centre for Causal Discovery (2023) to test FCI, GFCI and FGES, the Bayesys v3.5 package by Constantinou (2019) to test MAHC and SaiyanH, the GOBNILP v1.6.3 package by Cussens and Bartlett (2015) to test GOBNILP, the BiDAG package by Suter et al. (2023) to test Partition-MCMC, and the gCastle package by Zhang et al. (2021) to test NOTEARS-MLP. All algorithms are tested with their hyperparameter defaults as implemented in each package. Note that we also tested LiNGAM (Shimizu et al., 2006) that estimates the structure of linear causal models using the Tetrad package, as well as DAG-GNN (Yu et al., 2019) and MCSL (Ng et al., 2022) that perform nonlinear continuous optimisation using the gCastle package. However, whereas the other algorithms tested in this study completed structure learning within seconds or a few minutes, these three algorithms did not return a result within three hours of structure learning runtime and so they are not included in the results.

---

[4] A $v$-structure refers to the causal class of common-effect; i.e., $B \to A \leftarrow C$, where $A$ is common effect of $B$ and $C$.

Moreover, while this study considers a wide range of structure learning algorithms, there are other algorithms that we did not look at but are worth considering in future works. These include the score-based OBS (Teyssier and Koller, 2005) and its variants ASOBS (Scanagatta et al. 2015), INOBS (Lee and van Beek 2017) and WINASOBS (Scanagatta et al., 2017) that traverse the search-space of DAGs over different node orderings, where each ordering provides a unique set of constraints with regards to the orientation of edges, enabling structure learning with much larger datasets. When it comes to constraint-based learning, other algorithms include the modified version of PC by Li et al. (2019) that enforces the consistency of the separating sets of discarded edges with respect to the final graph, and Dual-PC by Giudice et al. (2022) that reduces the computational complexity of PC by changing the order in which the conditional independence tests are executed and prioritising certain high-order partial correlations.

*5.1. Model-averaging*

In addition to investigating the graphs learnt by the algorithms independently, we use a model-averaging procedure to obtain and assess graphs that are produced by a specific group of graphical outputs:

- **All_score-based:** The average graph over all score-based algorithms;
- **All_constraint-based:** The average graph over all constraint-based algorithms;
- **All_hybrid:** The average graph over all hybrid learning algorithms;
- **All_quartiles:** The average graph over all algorithms applied to data discretised using quartiles;
- **All_k-means:** The average graph over all algorithms applied to data discretised using k-means clustering;
- **All_continuous:** The average graph over all algorithms applied to continuous data;
- **All_mixed:** The average graph over all algorithms applied to mixed data.

We make available this model-averaging approach in the Bayesys package mentioned above. The input is a set of edges obtained from the multiple graphs. The model-averaging procedure prioritises directed edges over undirected edges under the assumption that a directed edge carries higher certainty than an undirected edge. The model-averaging procedure aims to orientate as many edges as possible (in our experiments, it orientated all edges). Bi-directed edges are ignored since they indicate incorrect dependence due to confounding, similar to how the absence of an edge indicates independence. Given a set of directed and undirected edges, including duplicate edges corresponding to multiple graphs, the model-averaging procedure works as follows:

1. Add <u>directed</u> edges to the average graph, starting from highest occurrence;
    a. Skip edge if already added in reverse direction;
    b. Skip edge if it produces a cycle, reverse it and add it to edge-set $C$;
2. Add <u>undirected</u> edges starting from highest occurrence;
    a. Skip edge if already added as directed;
3. Add <u>directed</u> edges found in $C$ starting from highest occurrence;
    a. Skip edge if already added as undirected.

An optional hyperparameter enables the user to specify the minimum number of occurrences needed for an edge to be considered in the average graph. We specify a threshold $\theta$ so that at least $\frac{1}{3}$ of the relevant learnt graphs included the edge. Specifically, :

- **All_score-based:** Was constructed over 31 independent graphs, and so $\theta = 10$;
- **All_constraint-based:** Was constructed over 4 independent graphs, and so $\theta = 2$;
- **All_hybrid:** Was constructed over 17 independent graphs, and so $\theta = 6$;
- **All_quartiles:** Was constructed over 19 independent graphs, and so $\theta = 7$;
- **All_k-means:** Was constructed over 19 independent graphs, and so $\theta = 7$;
- **All_continuous:** Was constructed over 14 independent graphs, and so $\theta = 5$;
- **All_mixed:** Was constructed over 5 independent graphs, and so $\theta = 2$;

**Table 2.** The 64 structure learning experiments. Information about the independent GitHub libraries can be found in Section 5.

| Exp | Algorithm | Package | Learning class | Output | Data format | Missingness |
|---|---|---|---|---|---|---|
| 1 | CCHM | Independent package | Hybrid | MAG | Continuous | Imputed |
| 2 | FCI | TETRAD | Constraint-based | PAG | Discrete (quartiles) | Imputed |
| 3 | FCI | TETRAD | Constraint-based | PAG | Discrete (k-means) | Imputed |
| 4 | FCI | TETRAD | Constraint-based | PAG | Continuous | Imputed |
| 5 | FGES | TETRAD | Score-based | CPDAG | Discrete (quartiles) | Imputed |
| 6 | FGES | TETRAD | Score-based | CPDAG | Discrete (k-means) | Imputed |
| 7 | FGES | TETRAD | Score-based | CPDAG | Continuous | Imputed |
| 8 | FGES | TETRAD | Score-based | CPDAG | Mixed | Imputed |
| 9 | GFCI | TETRAD | Hybrid | PAG | Discrete (quartiles) | Imputed |
| 10 | GFCI | TETRAD | Hybrid | PAG | Discrete (k-means) | Imputed |
| 11 | GFCI | TETRAD | Hybrid | PAG | Continuous | Imputed |
| 12 | GOBNILP | GOBNILP | Score-based (exact) | CPDAG | Discrete (quartiles) | Imputed |
| 13 | GOBNILP | GOBNILP | Score-based (exact) | CPDAG | Discrete (k-means) | Imputed |
| 14 | GOBNILP | GOBNILP | Score-based (exact) | CPDAG | Continuous | Imputed |
| 15 | HC | bnlearn | Score-based | CPDAG | Discrete (quartiles) | Imputed |
| 16 | HC | bnlearn | Score-based | CPDAG | Discrete (k-means) | Imputed |
| 17 | HC | bnlearn | Score-based | CPDAG | Continuous | Imputed |
| 18 | HC | bnlearn | Score-based | CPDAG | Mixed | Imputed |
| 19 | HC-aIPW | Independent package | Score-based | CPDAG | Discrete (quartiles) | Incomplete |
| 20 | HC-aIPW | Independent package | Score-based | CPDAG | Discrete (k-means) | Incomplete |
| 21 | HC-aIPW | Independent package | Score-based | CPDAG | Continuous | Incomplete |
| 22 | HCLC-V | Independent package | Hybrid | DAG | Discrete (quartiles) | Imputed |
| 23 | HCLC-V | Independent package | Hybrid | DAG | Discrete (k-means) | Imputed |
| 24 | MAHC | Bayesys | Score-based | DAG | Discrete (quartiles) | Imputed |
| 25 | MAHC | Bayesys | Score-based | DAG | Discrete (k-means) | Imputed |
| 26 | MMHC | bnlearn | Hybrid | CPDAG | Discrete (quartiles) | Imputed |
| 27 | MMHC | bnlearn | Hybrid | CPDAG | Discrete (k-means) | Imputed |
| 28 | MMHC | bnlearn | Hybrid | CPDAG | Continuous | Imputed |
| 29 | MMHC | bnlearn | Hybrid | CPDAG | Mixed | Imputed |
| 30 | NOTEARS | Independent package | Score-based (cont. opt.) | DAG | Continuous | Imputed |
| 31 | NOTEARS-MLP | gCastle | Score-based (cont. opt.) | DAG | Continuous | Imputed |
| 32 | Partition-MCMC | BiDAG | Hybrid | DAG | Discrete (quartiles) | Imputed |
| 33 | Partition-MCMC | BiDAG | Hybrid | DAG | Discrete (k-means) | Imputed |
| 34 | Partition-MCMC | BiDAG | Hybrid | DAG | Continuous | Imputed |
| 35 | PC-Stable | bnlearn | Constraint-based | PDAG | Discrete (quartiles) | Imputed |
| 36 | PC-Stable | bnlearn | Constraint-based | PDAG | Discrete (k-means) | Imputed |
| 37 | PC-Stable | bnlearn | Constraint-based | PDAG | Continuous | Imputed |
| 38 | PC-Stable | bnlearn | Constraint-based | PDAG | Mixed | Imputed |
| 39 | SaiyanH | Bayesys | Hybrid | CPDAG | Discrete (quartiles) | Imputed |
| 40 | SaiyanH | Bayesys | Hybrid | CPDAG | Discrete (k-means) | Imputed |
| 41 | SED (HC) | Independent package | Score-based | CPDAG | Discrete (quartiles) | Imputed |
| 42 | SED (HC) | Independent package | Score-based | CPDAG | Discrete (k-means) | Imputed |
| 43 | SED (ILP) | Independent package | Score-based | CPDAG | Discrete (quartiles) | Imputed |
| 44 | SED (ILP) | Independent package | Score-based | CPDAG | Discrete (k-means) | Imputed |
| 45 | SED (MMHC) | Independent package | Hybrid | CPDAG | Discrete (quartiles) | Imputed |
| 46 | SED (MMHC) | Independent package | Hybrid | CPDAG | Discrete (k-means) | Imputed |
| 47 | SED (PC) | Independent package | Hybrid | CPDAG | Discrete (quartiles) | Imputed |
| 48 | SED (PC) | Independent package | Hybrid | CPDAG | Discrete (k-means) | Imputed |
| 49 | SED (TABU) | Independent package | Score-based | CPDAG | Discrete (quartiles) | Imputed |
| 50 | SED (TABU) | Independent package | Score-based | CPDAG | Discrete (k-means) | Imputed |
| 51 | Structural EM | bnlearn | Score-based | CPDAG | Discrete (quartiles) | Incomplete |
| 52 | Structural EM | bnlearn | Score-based | CPDAG | Discrete (k-means) | Incomplete |
| 53 | Structural EM | bnlearn | Score-based | CPDAG | Continuous | Incomplete |
| 54 | TABU | bnlearn | Score-based | CPDAG | Discrete (quartiles) | Imputed |
| 55 | TABU | bnlearn | Score-based | CPDAG | Discrete (k-means) | Imputed |
| 56 | TABU | bnlearn | Score-based | CPDAG | Continuous | Imputed |
| 57 | TABU | bnlearn | Score-based | CPDAG | Mixed | Imputed |
| 58 | All_constraint-based | Bayesys | Model-averaging | DAG | n/a | n/a |
| 59 | All_continuous | Bayesys | Model-averaging | DAG | n/a | n/a |
| 60 | All_hybrid | Bayesys | Model-averaging | DAG | n/a | n/a |
| 61 | All_k-means | Bayesys | Model-averaging | DAG | n/a | n/a |
| 62 | All_mixed | Bayesys | Model-averaging | DAG | n/a | n/a |
| 63 | All_quartiles | Bayesys | Model-averaging | DAG | n/a | n/a |
| 64 | All_score-based | Bayesys | Model-averaging | DAG | n/a | n/a |

## 6    Evaluation and Results

We start by investigating the dimensionality of the models obtained from the graphs generated by the structure learning algorithms, the average graphs, and the graph constructed by human knowledge. We then evaluate these graphs in terms of graphical differences in subsection 6.2, in terms of inference and predictive validation in subsection 6.3, in terms of simulating the effect of hypothetical interventions in subsection 6.4, and in terms of sensitivity analysis in subsection 6.5. In subsection 6.6 we explore what some of these algorithms tell us about confounding. Lastly, we provide a qualitative evaluation of the graphical structures in subsection 6.7, with reference to plausible real-world relationships in the COVID-19 case study.

It is important to clarify that the aim is not to assess the algorithms independently, but rather to investigate the capabilities of structure learning in general, including exploring the validity of model-averaging described in Section 5. On this basis, we place a greater focus on what we can learn from structure learning algorithms collectively, rather than independently. As shown in the subsections that follow, the average graphs that tend to perform better than others across the different evaluation criteria investigated, are the *All_score-based* and *All_k-means*, which we present in Figure 2 and Figure 3. As we discuss in the subsections that follow, most of the independent graphs learnt by each algorithm, as well as the average graphs, tend to be dissimilar. The remaining five average graphs can be found in Appendix B, Figures B1, B2, B3, B4, and B5.

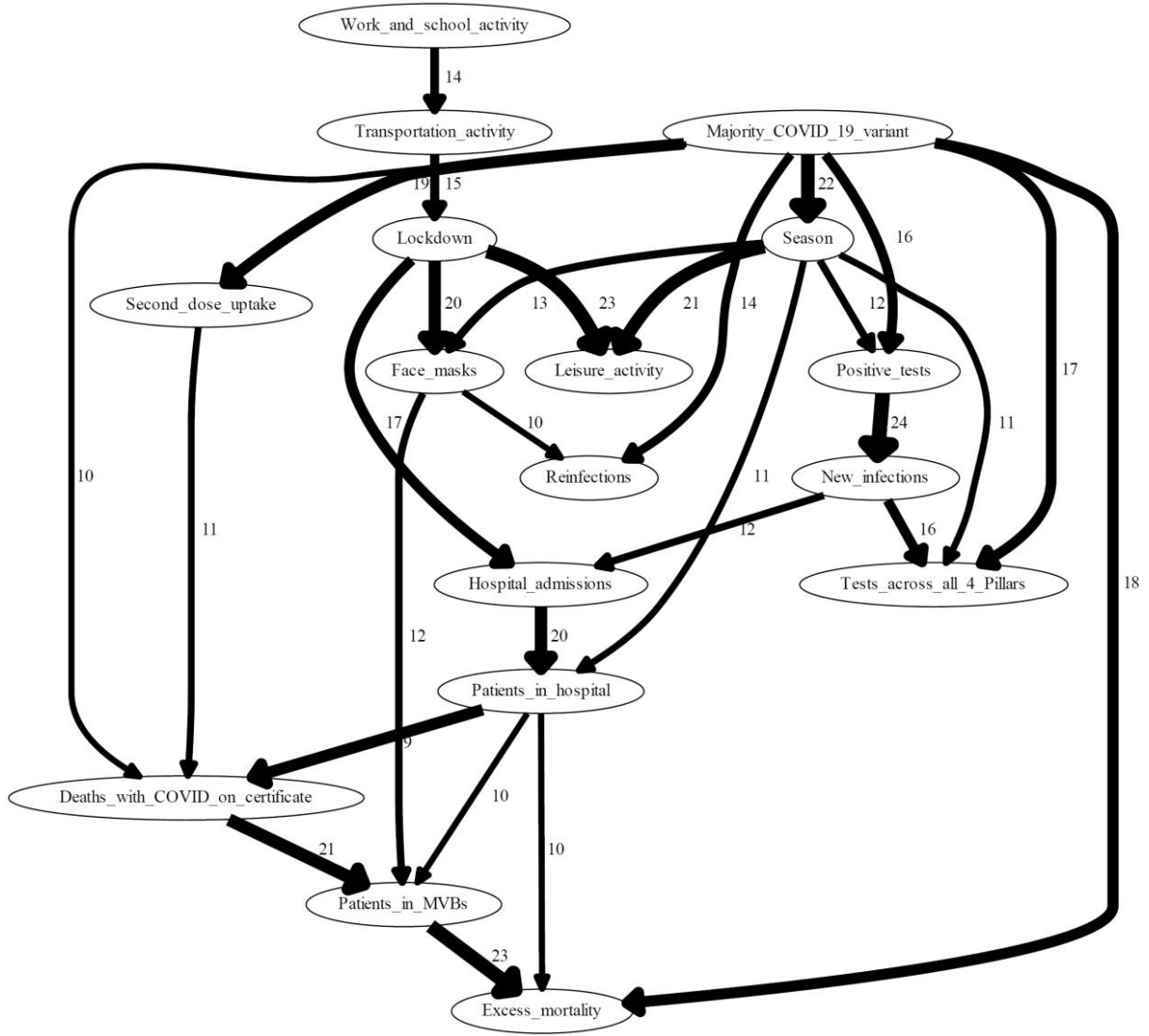

**Figure 2.** The *All_score-based* average graph obtained across 31 score-based experiments. The graph contains a total of 29 edges, where the edge labels represent the number of times the given edge appeared in the 31 outputs considered, and the width of the edges increases with this number. Edges that appeared less than 10 times across the 31 input graphs are not included.

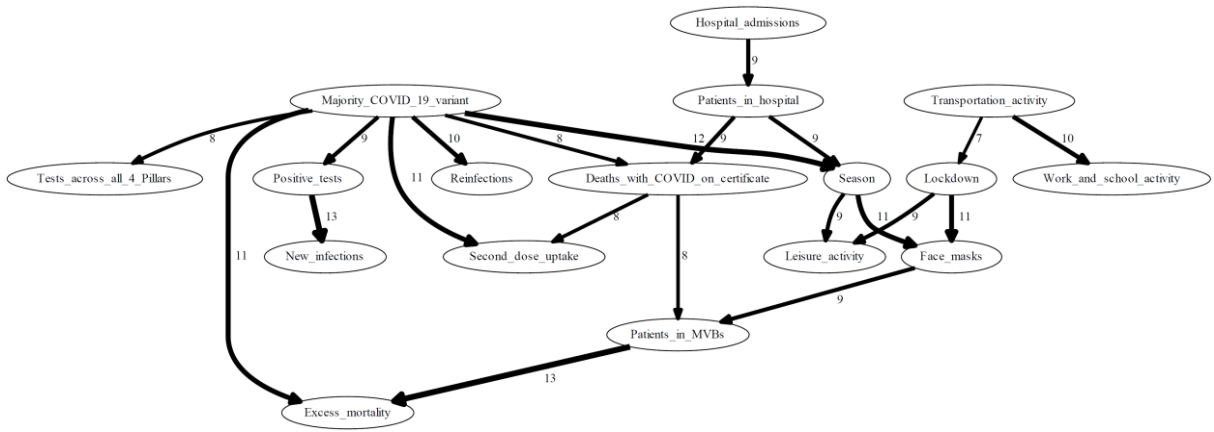

**Figure 3.** The *All_k-means* average graph obtained across 19 experiments where the input data set was discretised using k-means clustering. The graph contains a total of 21 edges, where the edge labels represent the number of times the given edge appeared in the 19 outputs considered, and the width of the edges increases with this number. Edges that appeared less than 7 times across the 19 input graphs are not included.

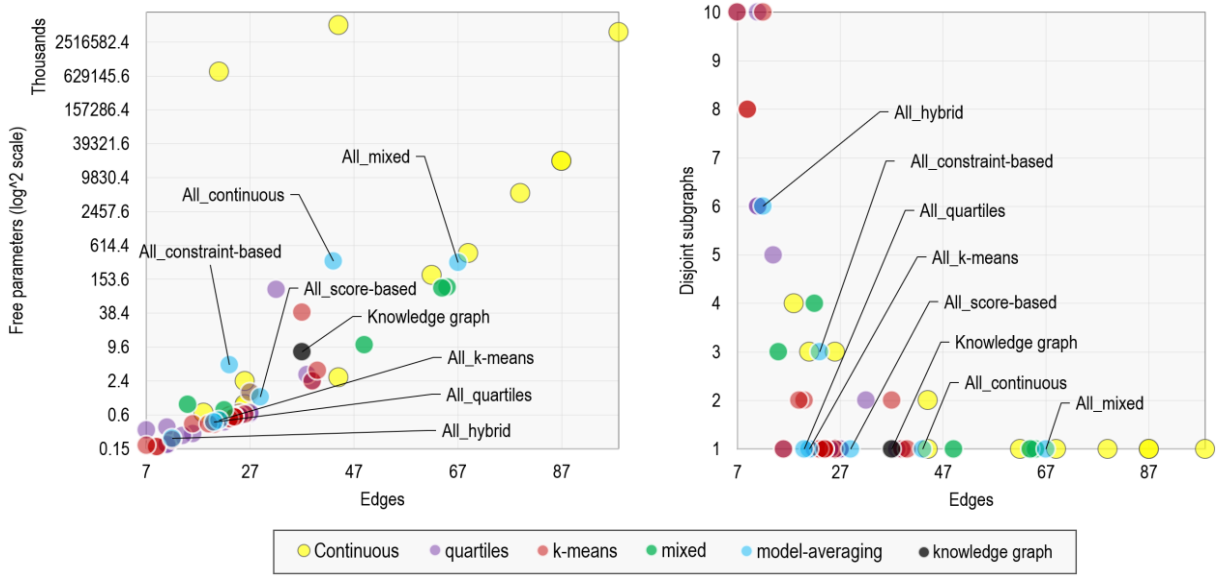

**Figure 4.** A two-dimensional scatter plot of the discrete models learnt by the independent algorithms, the average graphs, and the knowledge graph, classified by input data format.

*6.1. Dimensionality*

We start by investigating the dimensionality of the models obtained from the different graphical structures. While it is not always clear what level of model dimensionality might be best, the findings reported in this subsection are important in helping us understand the differences in graphical performance and inference reported in the subsequent subsections.

Figure 4 plots the number of edges against the number of disjoint subgraphs and the number of free parameters obtained by converting all graphs, irrespective of input data format, into discrete BN models. The number of free parameters, also known as independent parameters, is a measure of discrete model dimensionality. While it may not be appropriate to judge the structures recovered from continuous data in terms of free parameters, since this measure reflects the dimensionality of categorical or discrete distributions, we do this purely for comparison purposes to highlight the important differences in graph complexity. Further, the number of disjoint subgraphs represents the number of separate graphical fragments in a graph. For example, the average graph depicted in Figure B2 in Appendix B consists of six disjoint subgraphs. While, in theory, there is nothing wrong with an algorithm that produces graphs containing disjoint subgraphs, in practice we may require models that enable full propagation of evidence (Constantinou, 2020).

The scatterplots in Figure 4 quickly reveal that there are considerable differences between graphs learnt from data. Specifically, the number of edges in these graphs ranges from as low as 7 to as high as 98. The difference in the number of edges produced by the different algorithms and model-averaging approaches has a significant impact on the number of free parameters, which range from just 162 to above 5 billion. Note that this extreme difference is primarily due to the algorithms trained on continuous data, which – according to the results shown in Figure 4 – tend to produce highly dense graphs relative to those learnt from categorical data. This difference also applies to the same algorithms trained on different types of data.

It is important to clarify that some of the learnt structures lead to an unmanageable number of free parameters, in terms of inference, primarily due to the high number of parents they produce for some nodes, rather than due to the total number of edges they contain. For example, NOTEARS which led to the highest number of free parameters learnt just 44 edges, which is considerably fewer than many other algorithms, all of which led to lower dimensionality. Specifically, 15 out of the 44 edges learnt by NOTEARS are edges from 15 nodes entering a single node, which is rather unrealistic given that the data set contains just 17 variables; i.e., in the above case, the algorithm identified 15 parents out of possible 16 parents. Structural EM, which produced close to 4 billion free parameters, also produced 15 edges entering a single node; but in this case there were 98 edges in the learnt graph. NOTEARS-MLP produced around 1 billion free parameters, with a total of 14 edges entering a single node, out of the 21 edges discovered. These three extreme cases are followed by TABU and HC whose networks produced close to 20 million free parameters; i.e., 258 times fewer free parameters compared to those produced by NOTEARS and yet, still very high in terms of model dimensionality. Both TABU and HC learnt 87 edges, and had a maximum node in-degree of 11. All the five above extreme cases involve structure learning with continuous data, which explains why the learning process overestimated categorical dimensionality. Specifically, the free parameters produced by HC, TABU and Structural EM ranged between 552 and 653 when trained on categorical data, down from 20 million to 4 billion when trained on continuous data. This highlights that the data format has a considerable impact on the learnt output.

To summarise, the results in Figure 4 reveal a crucial difference between algorithms trained on data containing continuous variables (those categorised as *continuous* and *mixed* in Figure 4) and algorithms trained on discrete variables (those categorised as *quartiles* and *k-means*). That is, the former leads to denser graphs with possibly unmanageable dimensionality for inference when the learnt graphs are converted into discrete BNs, whereas the latter leads to sparser graphs that associate with lower model dimensionality. Note that this observation also applies to the same algorithms trained on different data formats. Unsurprisingly the knowledge graph is placed around the midpoint of the two extremes. In terms of disjoint subgraphs, the results follow a similar pattern in that the mixed and continuous data sets lead to graphs containing a lower number of disjoint subgraphs, whereas some of the sparser graphs learnt from discrete data contain up to 10 disjoint subgraphs.

The average graphs do well at softening the extreme differences. For example, the *All_continuous* graph contains fewer edges than those discovered by most of the algorithms that were trained on continuous data, whereas the *All_k-means* contains more edges than those discovered by most of the algorithms trained on categorical data discretised with k-means. The model-averaging approach also seems to lead to more realistic graphs in terms of model dimensionality and number of disjoint subgraphs.

*6.2. Graphical evaluation*

Graphical evaluation refers to the process of investigating graphical differences only, without accounting for inference or model fitting. In this subsection we investigate the graphical differences between the graphs learnt from data and the graph constructed using human knowledge.

We consider three relevant metrics. First, the Structural Hamming Distance (SHD) which corresponds to the number of changes needed to convert the learnt graph into the knowledge graph (or ground truth when this is known). Second, the F1 score which returns the harmonic mean of Recall ($R$) and Precision ($P$); i.e., $F1 = 2\frac{RP}{R+P}$. Third, the Balanced Scoring Function (BSF) which balances the score such that it considers the difficulty of discovering the presence of an edge versus the difficulty of discovering the absence of an edge:

$$BSF = 0.5 \left( \frac{TP}{a} + \frac{TN}{i} - \frac{FP}{i} - \frac{FN}{a} \right)$$

where $TP$, $TN$, $FP$ and $FN$ are the confusion matrix terms, $a$ is the numbers of edges and $i$ is the number of independencies in the ground truth, where $i = \frac{|V|(|V|)-1}{2} - a$ and $|V|$ is the number of nodes. The BSF score ranges from -1 to 1, where -1 corresponds to the worst possible graph; i.e., the graph that has edges present and edges absent between pairs of nodes that have edges absent and edges present respectively in the assumed ground truth. A score of 1 corresponds to the graph that matches the ground truth, and a score of 0 corresponds to a graph that is as informative as an empty or a fully connected graph. Lastly, these three metrics are applied to the corresponding CPDAG and PAG graphs; i.e., they represent the difference between Markov equivalence classes. The rules we have used to generate the scores are depicted in Table 3.

**Table 3.** The penalty weights (i.e., false positives or false negatives) used by the graphical evaluation metrics when comparing the learnt CPDAGs or PAGs to the knowledge graph. For PAG outputs containing edges o→ and o—o, we assume that they correspond to directed and undirected edges respectively for simplicity. These rules are taken from (Constantinou et al., 2022).

| Rule | True graph | Learnt graph | Penalty | Reasoning |
|---|---|---|---|---|
| 1 | A → B | A → B | 0 | Complete match |
| 2 | A → B | A ↔ B, A − B , A ← B | 0.5 | Partial match |
| 3 | any edge | no edge | 1 | No match |
| 4 | A ↔ B | A ↔ B | 0 | Complete match |
| 5 | A ↔ B | A − B , A ← B, A → B | 0.5 | Partial match |
| 6 | A − B | A − B | 0 | Complete match |
| 7 | A − B | A ↔ B , A ← B, A → B | 0.5 | Partial match |
| 8 | no edge | no edge | 0 | Complete match |
| 9 | no edge | Any edge/arc | 1 | No match |

Figure 5 presents a scatter plot that maps the SHD scores against the average of the F1 and BSF scores. The shaded quadrants separate results by median values. Specifically, outputs that fall within the green quadrant are those whose SHD and $\frac{F1+BSF}{2}$ scores are both better than median, whereas outputs that fall within the red quadrant are both below median. The yellow quadrants represent the case where the score is above median for one metric and below for the other. Some results of interest include:

i. The algorithms trained on continuous and mixed[5] data tend to fall within the red quadrant;
ii. The algorithms trained on categorical data tend to spread across all quadrants, irrespective of the approach to discretisation;
iii. Exact learning falls within the yellow and red quadrants;
iv. The graphs obtained from model-averaging tend to improve both scores, since they generally move towards the green quadrant.

One could argue that there is a strong disagreement between the knowledge graph and the graphs learnt from data, on the basis that no algorithm or average graph produced an overall $\frac{F1+BSF}{2}$ score greater than 0.35. The same can be concluded by looking at the SHD scores, where they range between 30 and 90. This is despite the knowledge graph presented in Figure 1 containing 37 edges only; implying that an empty graph would generate an SHD score of 37, which would be better than the score produced by most graphs learnt by the structure learning algorithms.

---

[5] Note that the algorithms applied to mixed data inherently impose a constraint that prohibits continuous variables to be parents of discrete variables. This is why in Figure B1, all four categorical variables *Season*, *Face_masks*, *Majority_COVID_19_variant*, and *Lockdown* have no continuous variables as parents.

arXiv pre-print, 2023.

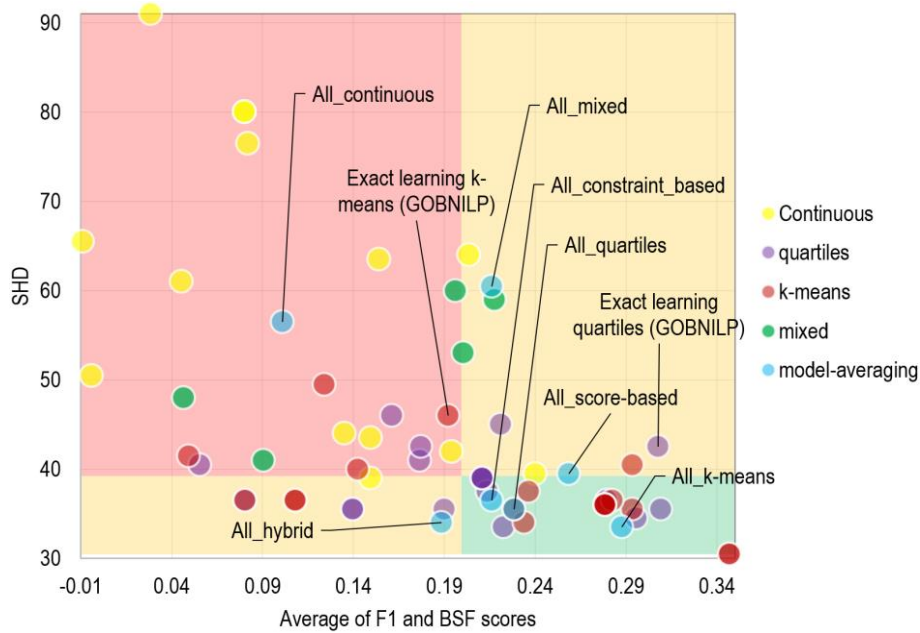

**Figure 5.** A two-dimensional scatter plot that presents the structural differences between the knowledge graph and the graphs produced by a) the independent algorithms classified by input data format, and b) the model-averaging approach applied to different groups of algorithms. The shaded quadrants represent median boundaries. A *lower* SHD score represents better performance, whereas a *higher* F1 and BSF score represents better performance.

We investigate these results further by measuring the similarity between average graphs and the knowledge graph. We present these results in Table 4, but we omit the BSF metric here since it is asymmetric; i.e., it is sensitive to the graph that we indicate as ground truth. Overall, the results show that most, but not all, scores between average graphs are closer than between the average graph and the knowledge graph. This means that while most algorithms generate graphs that are closer to graphs generated by some other algorithm, rather than the knowledge graph, some graphs generated from data have more differences between them compared to the differences they have with the knowledge graph. For example, the most similar average graphs require 10 graphical modifications (i.e., SHD is 10) to become identical, and the least similar require 62 graphical modifications to become identical.

**Table 4.** The F1 (below diagonal) and SHD (top diagonal) scores produced between different pairs of average graphs, and the knowledge graph. Green and red colours indicate stronger and weaker similarity respectively.

| Learning method | Knowledge | All_constraint-based | All_continuous | All_hybrid | All_k-means | All_mixed | All_quartiles | All_score-based |
|---|---|---|---|---|---|---|---|---|
| Knowledge |  | 36.5 | 56.5 | 33 | 32.5 | 59 | 34.5 | 39.5 |
| All_constraint-based | 0.283 |  | 34 | 17.5 | 25 | 57.5 | 23 | 30.5 |
| All_continuous | 0.238 | 0.424 |  | 33.5 | 38 | 64 | 35 | 38 |
| All_hybrid | 0.286 | 0.429 | 0.382 |  | 10.5 | 58 | 11 | 18 |
| All_k-means | 0.397 | 0.364 | 0.375 | 0.636 |  | 56.5 | 10.5 | 10 |
| All_mixed | 0.385 | 0.3 | 0.345 | 0.253 | 0.33 |  | 51.5 | 50 |
| All_quartiles | 0.333 | 0.419 | 0.413 | 0.625 | 0.707 | 0.379 |  | 15.5 |
| All_score-based | 0.348 | 0.327 | 0.417 | 0.537 | 0.76 | 0.438 | 0.633 |  |

*6.3. Inference-based and predictive validation*

Recall that the graphical evaluation presented in subsection 6.2 considers graphical structure only, and does not account for the inference capabilities of the models. This subsection explores the inference capabilities that arise by parameterising the learnt structures into discrete BNs. We assess the learnt BN in terms of model fitting, model selection, and cross-validation tests applied to categorical data; i.e., on the data sets discretised with k-means and quartile discretisation methods.

We start by looking at the Log-Likelihood ($LL$) and Bayesian Information Criterion ($BIC$) scores, where the former represents how well the model fits the data and the latter is a model selection function that balances model dimensionality with model fitting. Figure 6 plots these scores on two-dimensional scatter plots; one for each data discretisation case. The plot on the left is based on data discretised with quartiles and the plot on the right is based on data discretised with k-means clustering.

The results in Figure 6 show that, overall, the models learnt from graphs derived from data discretised with k-means produce higher $LL$ fitting scores compared to those derived from data discretised using quartiles. This is expected since quartiles aim for balanced distributions, whereas a clustering method such as k-means would discover distributions that maximise fitting. Despite the difference in $LL$ fitting, the results between the two plots appear to be consistent.

We find that the *All_continuous* and *All_mixed* graphs produce far worst model selection scores, and this observation is consistent with the results in 6.1 that show that continuous variables lead to denser graphs that, when converted into categorical models, tend to contain enormous numbers of parameters, and so have large complexity penalties in model selection scores. We also note that the increase in model dimensionality is far too big to be justified by any improvements in model fitting. That is, the $BIC$ score produced by *All_mixed* is the second worse in Figure 6 despite having the highest $LL$ score, which suggests that the increase in $LL$ fitting is likely to be due to overfitting. Moreover, the *All_continuous* model has a lower $LL$ fitting score than *All_score-based* despite the former being a considerably simpler model (according to the $BIC$ scores).

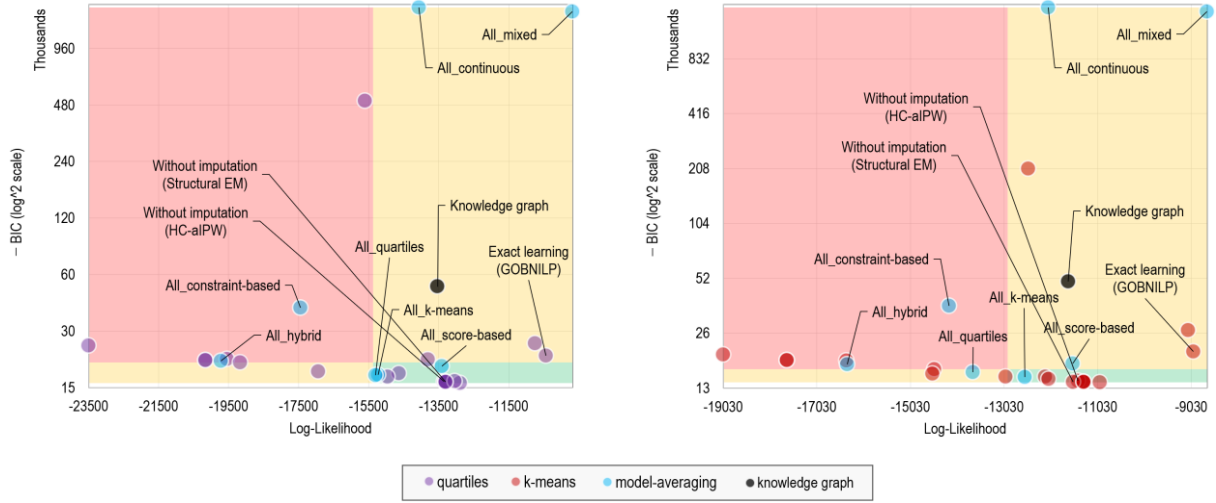

**Figure 6.** A two-dimensional scatter plot on the relationship between log-likelihood and the BIC score. The left plot assumes input data discretised with quartiles, and the right plot input data discretised with k-means clustering. The shaded quadrants represent median boundaries. Algorithms returning MAG or PAG outputs are excluded.

Other results of interest include the models obtained from the knowledge and the *All_constraint-based* graphs, which generate relatively poor model selection scores. In the case of the knowledge graph, this result is not surprising since we would *not* expect knowledge to outperform algorithms on what they are designed to do; i.e., typically maximise some model selection score or objective function. In the case of *All_constraint-based*, this observation is partly explained by the class of learning (i.e., constraint-based), which focuses on local conditional independence tests that tend to overlook global scores such as those derived from *LL* and *BIC*. This also explains why *All_score-based* fairs much better both in terms of *LL* and *BIC* scores. Lastly, and somewhat surprisingly, the exact learning GOBNILP produced a model selection score below median, despite designed to return the global maximum. A possible explanation is the hyperparameter restriction exact learning has on maximum node in-degree, which is '3' by default to handle computational efficiency, which means that graphs containing nodes with more than three parents, and which could have led to higher model selection scores, were not explored. This usually does not fully explain the relatively poor performance of exact learning GOBNILP on model selection score, however, which may be influenced by data noise that is naturally present in real data. For example, in Constantinou et al. (2021) it was shown that simple learners are more resilient to data noise compared to exact or more sophisticated approximate learners.

In addition to the model fitting and model selection scores, we also investigate the inference capability of the learnt models in terms of cross-validation performance. We load the learnt models into the GeNIe BN software (BayesFusion, 2022) and run a 10-fold cross-validation on the two discrete data sets. Figure 7 presents the average cross-validation classification accuracy achieved over all 17 variables; i.e., the average of 17 runs of 10-fold cross-validation for predicting each of the 17 variables. Note that a 1% discrepancy in cross-validation accuracy represents a rather meaningful difference, since that would imply a 1% difference in classification accuracy as the average across all 17 nodes.

The results in Figure 7 show some consistencies with previous results. Specifically, structure learning with continuous variables is, once more, found to lead to poor performance. This reinforces our hypothesis that the increase in graph density observed in graphs learnt from continuous variables is likely due to model overfitting. Moreover, some of the graphs learnt from continuous data could not have been parameterised due to extremely high dimensionality (refer to Figure 4), and these cases are not included in Figure 7. These cases involve the continuous data versions of HC, TABU, Structural EM, NOTEARS and NOTEARS-MLP which could not be parameterised, and PC (all data versions) whose PDAG outputs cannot always be extended into a DAG structure. On the other hand, many of the structures learnt from mixed data did produce strong performance, and this somewhat contradicts previous results on continuous data. Interestingly, the average graphs *All_k-means* and *All_score-based* have performed well in terms of cross-validation too, especially under the k-means case.

*6.4. Interventional analysis*

Causal modelling enables simulation of intervention. We use the GeNIe software (BayesFusion, 2022) that implements Pearl's *do*-operator, which is a mathematical representation of an intervention (Pearl, 2012), to measure the effect of hypothetical interventions and compare these effects across different learnt structures. We perform interventional analysis on a set of variables of interest that would be reasonable in real-life for intervention, given the COVID-19 case study. However, recall that because policy interventions captured by data reflect observational relationships without time shift (e.g., current policy relates to current status of pandemic), we cannot simulate the effect of policy interventions since that would require that we test for distribution shifts (e.g., exploring the future effect of policy) – which is something these algorithms do not explore; we discuss this limitation in detail in Concluding remarks. Therefore, we measure the impact of hypothetical intervention on the following:

1. The number of COVID-19 tests; i.e., *do(Tests_across_all_4_Pillars)* as shown in Table 6. For example, we may be interested in measuring whether there is any benefit in increasing the number of COVID-19 tests.
2. The number of COVID-19 infections; i.e., *do(New_infections)*. For example, we may be interested in measuring the effect of controlling the rate of infection.
3. The number of COVID-19 patients in hospital; i.e., *do(Patients_in_hospital)*. For example, to explore the benefit of reducing the risk of hospitalisation or discharging hospital patients earlier.
4. The number of COVID-19 patients in MVBs; i.e., *do(Patients_in_MVBs)*. For example, to explore the benefit of medication or treatment that may reduce the risk of requiring a MVB.

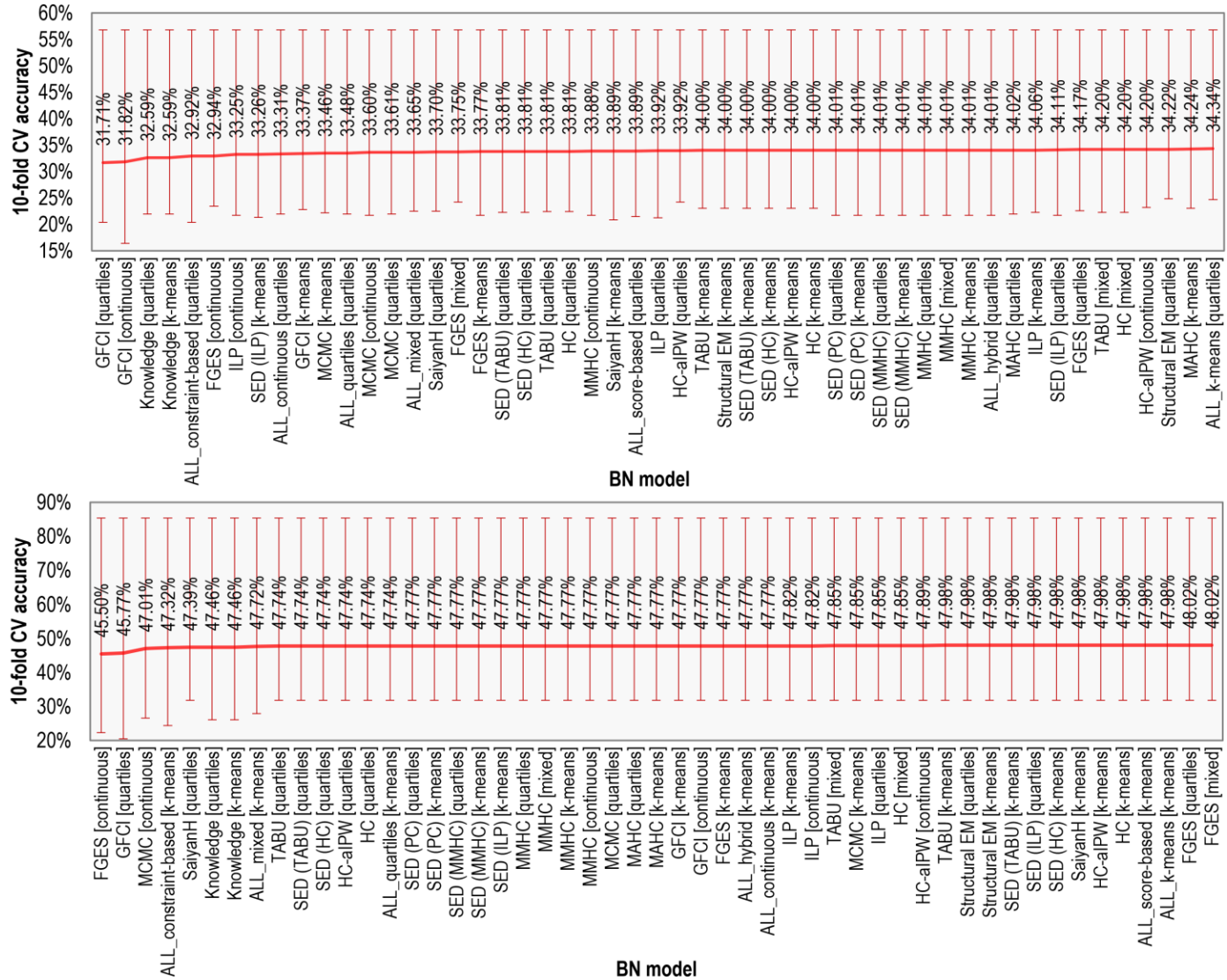

**Figure 7.** 10-fold cross-validation classification accuracy averaged across all the 17 nodes, ordered by worse to best performance. The chart at the top is based on models parameterised with categorical data discretised using quartiles, whereas the chart at the bottom is based on categorical data discretised using k-means clustering. Algorithms returning MAG or PAG outputs that included bi-directed edges are excluded from this assessment, as well as some graphs learnt by HC, TABU, Structural EM, NOTEARS, NOTEARS-MLP, and PC (refer to main text for explanation). The width of the error bars represents the distance from the lowest and highest CV scores across all 17 nodes.

Table 5 describes the linear scale we use to measure the effect of intervention, and Table 6 presents the effect of the above interventions on a set of four variables of interest, given each of the seven average graphs and the knowledge graph. Because all the variables are represented by an ordinal distribution, we measure the effect of intervention in terms of distribution shift, and we normalise this shift on a linear scale between 0 and 1. Consider that we are intervening on variable $A$ and subsequently measure the effect of intervention on variable $B$, then (and assuming deterministic distributions for simplicity):

**Table 5.** Measuring the effect of hypothetical intervention. The effect score is weighted[6] given the probabilities assigned to each state.

|  | $do$ ($A$=Very_Low) | $do$ ($A$=Very_High) | Effect score |
|---|---|---|---|
| Effect on distribution $B$ | $p$(Very_Low) = 1 | $p$(Very_High) = 1 | 1 |
|  | $p$(Very_High) = 1 | $p$(Very_Low) = 1 | 1 |
|  | $p$(Very_Low) = 1 | $p$(High) = 1 | 0.66 |
|  | $p$(Low) = 1 | $p$(Very_High) = 1 | 0.66 |
|  | $p$(Very_High) = 1 | $p$(Low) = 1 | 0.66 |
|  | $p$(High) = 1 | $p$(Very_Low) = 1 | 0.66 |
|  | $p$(Very_Low) = 1 | $p$(Low) = 1 | 0.33 |
|  | $p$(Low) = 1 | $p$(High) = 1 | 0.33 |
|  | $p$(High) = 1 | $p$(Very_High) = 1 | 0.33 |
|  | $p$(Low) = 1 | $p$(Very_Low) = 1 | 0.33 |
|  | $p$(High) = 1 | $p$(Low) = 1 | 0.33 |
|  | $p$(Very_High) = 1 | $p$(High) = 1 | 0.33 |
|  | $p$(Very_Low) = 1 | $p$(Very_Low) = 1 | 0 |
|  | $p$(Low) = 1 | $p$(Low) = 1 | 0 |
|  | $p$(High) = 1 | $p$(High) = 1 | 0 |
|  | $p$(Very_High) = 1 | $p$(Very_High) = 1 | 0 |

The results presented in Table 6 show that the different structures can produce very different interventional effects. These results contrast the results from predictive validation in subsection 6.3, which show that large differences in graphical structure translate to small differences in predictive performance. Results worth highlighting include:

i. **Effect on testing:** Approximately half of the average graphs indicate that intervening on infections would moderately affect testing, and intervening on hospitalisations and MVBs would weakly affect testing. On the other hand, the knowledge graph - as well as the other half of the average graphs - show no effect on testing, under the assumption that it is the number of tests that determine the number of positive cases.
ii. **Effect on positive tests:** The knowledge graph suggests that intervening on the number of infections would greatly affect the number of positive tests observed. All the other interventions and average graphs suggest no to negligible impact on the number of positive tests.
iii. **Effect on infections:** All graphical structures agree that none of the interventions explored has a meaningful effect on the number of infections.
iv. **Effect on deaths:** All graphical structures agree that intervening on the number of tests and on the rate of infection, would (rather surprisingly) have no to negligible effect on COVID-19 deaths. On the other hand, the knowledge graph, including most of the average graphs, agree that intervening on hospitalisations would greatly affect the number of deaths. Interestingly, the knowledge graph is the only one that suggests that intervening on the probability patients would require admission to MVBs would have an impact on the number of deaths.

---

[6] For example, if $do(A=\text{Very\_Low})$ produces an effect distribution { 0.5, 0.25, 0.15, 0.1 } with score 0.5×0+0.25×0.33+0.15×0.66+0.1×1=0.2815, and $do(A=\text{Very\_High})$ an effect distribution {0.6, 0.3, 0.1, 0} with score 0.6×0+0.3×0.33+0.1×0.66+0×1=0.165, then that would produce an effect score of |0.2815-0.165|=0.1165.

**Table 6.** The effect of the specified hypothetical interventions on four variables of interest. The effect of intervention is presented in terms of distribution shift normalised between 0 and 1 as described in Table 5. The results are distributed across the seven average graphs and the knowledge graph. No result/red bar indicates no effect.

| | | Tests_across_all_4_Pillars | Positive_tests | New_infections | Deaths_with_COVID_on_certificate |
|---|---|---|---|---|---|
| do(Tests_across_all_4_Pillars) | Knowledge graph | | 0.033 | | 0.003 |
| | ALL_constraint-based | | | | |
| | ALL_continuous | | | | |
| | ALL_hybrid | | | | |
| | ALL_k-means | | | | |
| | ALL_mixed | | | | |
| | ALL_quartiles | | | | |
| | ALL_score-based | n/a | | | |
| do(New_infections) | Knowledge graph | | 0.428 | | 0.04 |
| | ALL_constraint-based | | | | |
| | ALL_continuous | 0.066 | | | |
| | ALL_hybrid | | | | |
| | ALL_k-means | | | | |
| | ALL_mixed | 0.209 | | | |
| | ALL_quartiles | 0.299 | | | |
| | ALL_score-based | 0.299 | | n/a | |
| do(Patients_in_hospital) | Knowledge graph | | | | 0.5599 |
| | ALL_constraint-based | | | | |
| | ALL_continuous | 0.0432 | 0.0397 | 0.0264 | |
| | ALL_hybrid | | | | |
| | ALL_k-means | | | | 0.5599 |
| | ALL_mixed | 0.0964 | | | 0.3088 |
| | ALL_quartiles | | 0.0198 | 0.0198 | 0.6648 |
| | ALL_score-based | | | | 0.4636 |
| do(Patients_in_MVBs) | Knowledge graph | | | | 0.3126 |
| | ALL_constraint-based | | | | |
| | ALL_continuous | 0.1062 | 0.0066 | | |
| | ALL_hybrid | | | | |
| | ALL_k-means | | | | |
| | ALL_mixed | 0.0963 | | | |
| | ALL_quartiles | | 0.0067 | 0.0067 | |
| | ALL_score-based | | | | |

*6.5. Sensitivity analysis*

Unlike interventional analysis which explores the effect of hypothetical interventions by looking at the impact interventions have on children and descendant nodes, sensitivity analysis looks at how sensitive a node is to its parent and ancestor nodes. Specifically, high sensitivity indicates that small changes to the CPT parameters of a node will have a strong effect on its posterior distributions, thereby indicating dependency between variables that are sensitive to one another given the BN structure, whereas low sensitivity signifies that large changes to the CPT parameters will have a weak effect on its posterior distributions, thereby indicating no dependency between nodes that are insensitive to one another. Sensitivity analysis was initially proposed by Castillo et al. (1997) as a method of evaluating the probabilistic parameters of BNs.

We use the GeNIe BN software (BayesFusion, 2022) to perform sensitivity analysis. For simplicity, we apply sensitivity analysis to a single variable, which illustrates some of the discrepancies in conclusions on sensitivity between the different learnt structures. We focus on the number of deaths due to COVID-19, under the assumption that this would be one of the key variables of interest.

Figure 8 presents the sensitivity of node *Deaths_with_COVID_on_certificate* as determined by the graph obtained from knowledge and the average graphs. We present only the relevant fragments of those graphs, containing the parent and ancestor nodes of the target node on which sensitivity analysis is applied. Nodes that are not in the set of parent and ancestor nodes of the target variable are coloured in grey, with the rest coloured in red where a darker red colour indicates stronger sensitivity. Note that comparisons between different shades of red colour should not be made across different graphs.

The inconsistencies in the results obtained from the different learnt structures extend to sensitivity analysis. Specifically, the *All_constraint-based* and *All_hybrid* graphs suggest that deaths due to COVID-19 are insensitive to all other nodes, since these graphs did not contain any parents of target node *Deaths_with_COVID_on_certificate*. On the other hand, the *All_continuous*, *All_k-means* and *All_quartiles* graphs suggest that deaths are most sensitive to hospitalisations, whereas the *All_score-based*, *All_k-means* and the knowledge graphs suggest that deaths are most sensitive to the COVID-19 variant. On the contrary, the *All_mixed* graph suggests that deaths from COVID-19 are most sensitive to lockdowns.

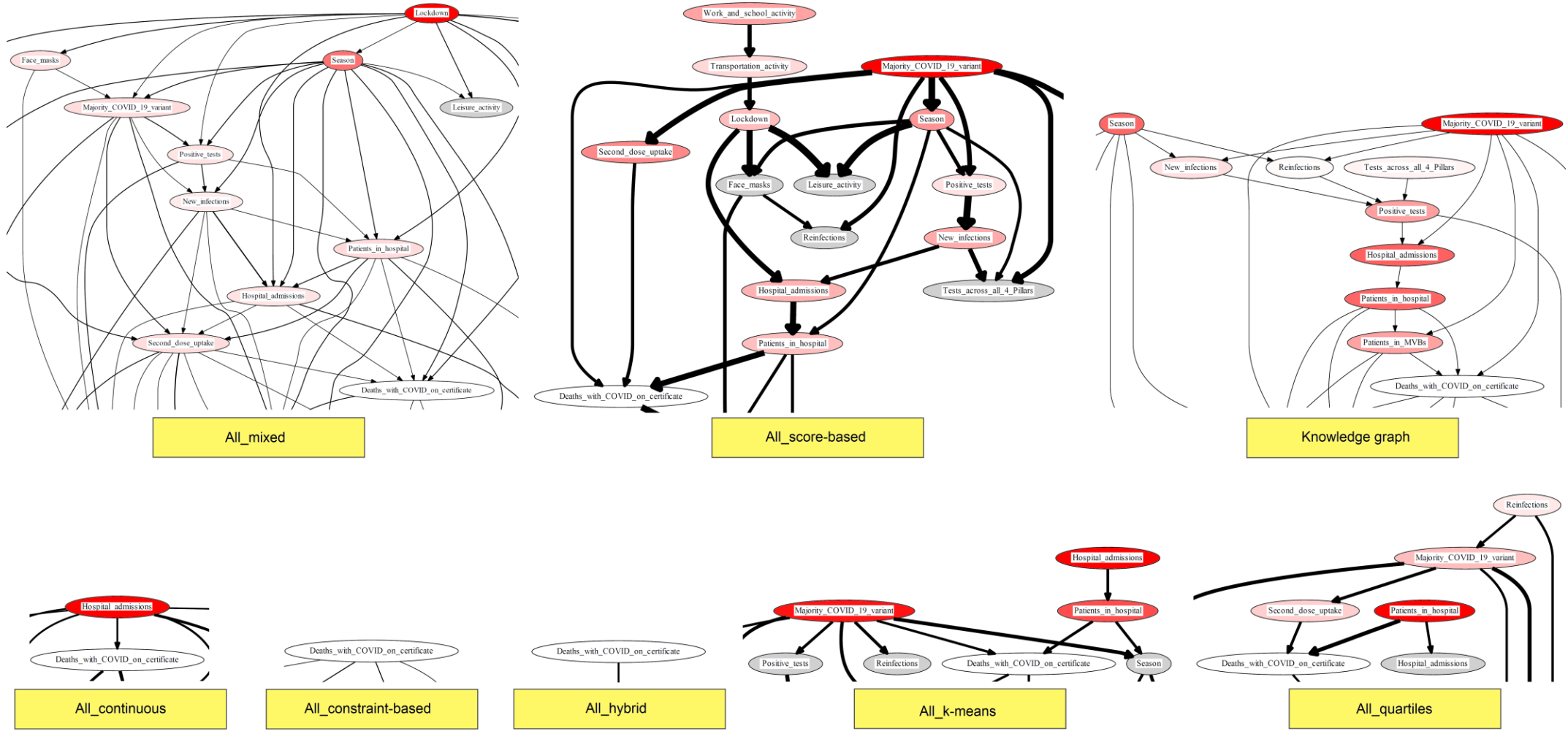

**Figure 8.** Sensitivity analysis on target node *Deaths_with_COVID_on_certificate* given the knowledge graph and the average graphs, trained on discrete k-means data. Only graphical fragments containing the relevant parent and ancestor nodes are shown (note the fragments preserve the edge-width from the average graphs). White nodes represent the target node, grey nodes are not in the set of parent and ancestor nodes, and red nodes indicate sensitivity to target node where a darker red colour implies stronger sensitivity.

arXiv pre-print, 2023.

*6.6. Confounding evaluation*

This subsection investigates the algorithms that assume causal insufficiency, which is the assumption that the input data do not contain all of the variables of interest, some of which could be missing common-causes (i.e., latent confounders) that confound the learnt structure. As shown in Table 1, nine of the structure learning experiments are based on algorithms that are capable of producing graphs containing bi-directed edges indicating confounding. These experiments involve the FCI, GFCI, CCHM and HCLC-V algorithms that generate a MAG or a PAG output.

Figure 9 presents the average graph that contains the bi-directed edges discovered (excludes any directed or undirected edges), indicating possible confounding between pairs of variables. This average graph was generated using the same model-averaging approach described in subsection 5.1, but with a focus on bi-directed edges instead. However, because just two bi-directed edges satisfied threshold $\theta$ as defined in Section 5 shown in black colour in Figure 9 (in this case $\theta = 3$), we chose to also present the edges that failed to meet this threshold by a single count, shown in red colour.

The results on confounding are also found to be inconsistent between the different algorithms and data formats considered. For example, across the nine experiments, only three of them contain the same two bi-directed edges shown in Figure 9, and this level of agreement decreases for all other edges. Moreover, given that a bi-directed edge indicates confounding, which in turn leads to spurious edges, it implies that algorithms that assume causal sufficiency (i.e., all the other algorithms investigated in this paper) would have discovered these spurious edges as true edges. In other words, we would expect the algorithms tested in the previous subsections to have produced an edge between the pairs connected with a bi-directed edge in Figure 9. However, with reference to the black coloured bi-directed edges in Figure 9, only four out of the seven average graphs contain an edge between *Excess_mortality* and *Patients_in_hospital*, and only two contain an edge between *Tests_across_all_4_Pillars* and *Work_and_school_activity*; and most of these edges are found in *All_continuous* and *All_mixed* graphs which are dense graphs. This observation highlights further disagreements between the different structure learning approaches investigated.

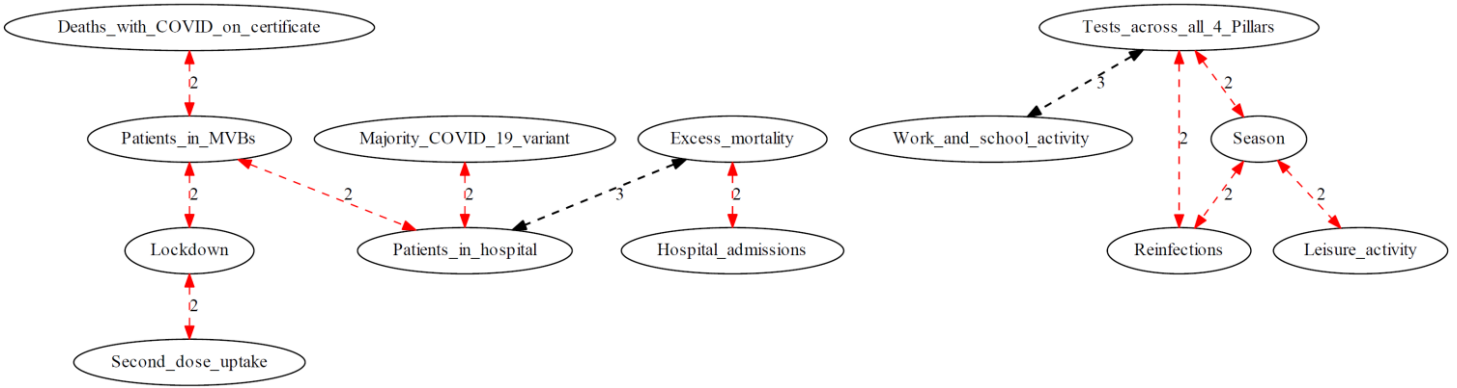

**Figure 9.** An average graph containing bi-directed edges only (excludes directed and undirected edges), indicating confounding. Edges in black colour represent those that satisfy threshold $\theta$ as defined in Section 5 with reference to the model-averaging process. Because only two edges satisfied this criterion (i.e., $\theta = 3$ given the nine individual graphs which are averaged), we also present the bi-directed edges that failed to meet this threshold by a single count, coloured in red.

*6.7. Case-study evaluation*

This subsection focuses on qualitative evaluation of the causal relationships recovered by the structure learning algorithms. We focus on relationships that are well-understood and for which we are relatively confident about their presence in the ground truth. Therefore, these qualitative assessments are based on a series of knowledge-based assumptions about what might be, and what might not be, correct. This subjective assessment is depicted in Table 7, and focuses on the seven average graphs.

**Table 7.** Qualitative evaluation of the seven average graphs with reference to the listed COVID-19 relationships that are assumed to be common knowledge. We assume the learnt outputs produced by structure learning are in agreement with causal knowledge if more than half of the learnt outputs are consistent with that knowledge.

| Qualitative evaluation | Expectation | Observation | In agreement with causal knowledge? |
|---|---|---|---|
| Infections | A relationship between infections (any) and testing (any), without specific requirement about the direction of causation. | Relationship present in 7 out of 7 graphs. | Yes |
| Vaccination | A relationship between *Second_dose_uptake* and *Deaths_with_COVID_on_certificate* or *Excess_mortality*, without specific requirement about the direction of causation. | Relationship present in 4 out of 7 graphs. | Yes |
| Testing | A relationship between testing (any) and infections (any), without specific requirement about the direction of causation. | Relationship present in 7 out of 7 graphs. | Yes |
| Deaths | A relationship between hospitalisations (any) and deaths from COVID-19. Causation must travel from hospitalisations to deaths. | Relationship present in 7 out of 7 graphs, and orientated correctly in 0 graphs. | Partly |
| Hospitalisations | A relationship between *Hospital_admissions* and *Patients_in_hospital* or *Patients_in_MVBs*. Causation must travel from hospital admissions to patients in hospital/MVBs. | Relationship present in 7 out of 7 graphs, and orientated correctly in 3 graphs. | Partly |
| Policy | A relationship between *Lockdown* and mobility (any). Causation must travel from lockdowns to mobility. | Relationship present in 7 out of 7 graphs, and orientated correctly in 2 graphs. | Partly |
| Policy | A relationship between *Lockdown* and hospitalisations (any). Causation must travel from hospitalisations to lockdowns. | Relationship present in 2 out of 7 graphs, and orientated correctly in 0 graphs. | No |

Despite the rather strong inconsistencies between learnt structures highlighted in the previous subsections, the results in Table 7 suggest that structure learning has performed well in identifying, or partly identifying, most of the relationships that we assume to represent common knowledge. The only relatively strong disagreement between learnt structures and common knowledge involves the relationship between hospitalisations and lockdowns. This specific relationship, however, involves a temporal aspect that could not have been identified by structure learning algorithms that assume that the conditional distributions remain static over time. More specifically, we know that lockdowns were imposed when hospitalisations were high, and lockdowns were relaxed when hospitalisations were low. However, just like most ML algorithms, because the structure learning algorithms investigated do not expect the relationships between variables to be time-varying, what these algorithms read from data is that hospitalisations were both low and high in both the presence and absence of lockdowns, and fail to identify the temporal trigger for lockdown or the temporal effect of lockdown; i.e., hospitalisations tend to decrease a couple of weeks following lockdown, depending on lockdown severity. Therefore, not finding a relationship between hospitalisations and lockdown highlights the inability of these algorithms to recover such time-varying relationships, rather than pointing to a significant finding.

## 7 Concluding remarks

This study investigates some of the open problems in causal structure learning as they apply to the modelling of the COVID-19 pandemic. We formulate 64 experiments that are based on 29 different algorithms and apply them to data we collated about the COVID-19 pandemic in the UK. Since causal models enable us to simulate the effect of hypothetical interventions, we assume that the problem of COVID-19, which required swift unprecedented decisions in response to previously unobserved events, serves as an excellent testbed for causal structure learning. We use this section to summarise open problems in causal structure learning based on this case study, and to formulate directions for future work. To facilitate future work, we make all the learnt graphs, average graphs, BN models, data sets, and source code publicly available online through the Bayesys repository[7] (Constantinou et al., 2020).

The process we followed to investigate the set of structure learning algorithms is exhaustive and can be applied to any dataset. Specifically,
   i. Subsection 3.2 describes the processes we followed to convert the raw mixed dataset into discrete datasets and into a continuous dataset. Additionally, the subsection also describes how we impute missing data values under the assumption the missing values are not missing at random.
   ii. Section 5 describes the set of algorithms considered, spanning different categories as well as different classes of structure learning, along with the packages or software used to apply each of these algorithms to data.
   iii. Subsection 5.1 describes the model-averaging approach we used to obtain average graphs from the outputs generated by different groups of structure learning algorithms,
   iv. Subsection 6.1 describes the functions and metrics we considered to measure the dimensionality of each learnt structure.
   v. Subsection 6.2 describes the graphical metrics used to measure the graphical differences between different learnt structures.
   vi. Subsection 6.3 describes the process we followed to assess each of the learnt structures in terms of inference capability. This process includes model-selection and goodness-of-fit functions, as well as predictive capability via cross-validation across all nodes contained within the learnt structures.
   vii. Subsection 6.4 describes how we simulate hypothetical interventions on variables of interest, to estimate the effect of these interventions as determined by the different learnt structures.
   viii. Subsection 6.5 describes how we perform sensitivity analysis between a given node and its parent and ancestor nodes, given the different learnt structures.
   ix. Subsection 6.6 describes the structure learning algorithms we have used to identify possible latent confounders that may confound the observed variables present in the input data.

The main findings of this study include a) learnt structures are found to be highly sensitive to the choice of the algorithm, b) model-averaging effectively and efficiently reduces algorithm-based sensitivity, c) structures learnt from continuous data are dense and likely prone to overfitting and spurious causal relationships compared to the structures learnt from discrete or categorical data, and d) structure learning not accounting for distribution shifts is a critical limitation. We discuss each of these findings in detail below.

The most evident outcome is that the learnt structures are found to be highly inconsistent across the various structure learning algorithms considered. That is, the algorithms produce graphs that are very different in the number of edges they contain, the actual edges discovered, and the orientation of those edges. This inconsistency increases when comparing algorithms from different learning classes (e.g. score-based or constraint-based), and by input data format (e.g., categorical or continuous). While one could argue that it is, to some extent, reasonable for algorithms that rely on different classes of learning to behave differently, the results show that these inconsistencies are largely present across algorithms of the same learning class and input data format. It is also important to highlight that this level of disagreement between algorithms leads to trivial differences in predictive validation (see Figure 7), but to considerable differences when the evaluation is extended to interventional (see Table 6) or sensitivity (see Figure 8) analyses. These empirical findings further highlight the inability of predictive validation in providing meaningful answers to questions about causal reasoning.

---

[7] http://constantinou.info/downloads/bayesys/bayesys_repository.pdf

Many of the structural discrepancies cannot be explained by differences between algorithms, since the same algorithm would often produce very different graphs depending on the input data format. A common approach towards reducing the inconsistency in the learnt graphs involves performing model-averaging across a set of graphs, to obtain an average graph that is representative of that set of learnt graphs. This is something that we also investigate, by grouping algorithms in terms of learning class or data format. While model-averaging is found to indeed reduce variability, we also find that the average graphs for each group are all different from one another. Future extensions of this work could focus on developing more sophisticated approaches to model-averaging. For instance, the current study assumes equal contribution from each structure learning algorithm in the average graph. However, it may be beneficial to assume a weighted average that prioritises edges learnt by algorithms known to be more accurate than others.

This level of inconsistency extends to algorithms that support latent variables (also known as algorithms that assume *causal insufficiency*). Because these algorithms aim to recover structures that highlight possible spurious relationships that are the result of latent confounders, we would expect these spurious edges to have been discovered as edges present in the learnt graphs by algorithms that do not account for latent confounders. However, our findings not only show that the algorithms that assume causal insufficiency recover contrasting spurious edges, but also that many of the predicted spurious edges are not present in most of the structures learnt by algorithms that do not support latent variables. These inconsistencies in confounding effects raise questions about the effectiveness of the structure learning algorithms that support latent variables. One possible future research direction would be to place a greater focus on improving and properly evaluating algorithms aimed at recovering latent confounders. For instance, it would be useful to know the success rate of these algorithms in correctly identifying latent confounders, by considering both the true positive and false positive rates. Additionally, it is important to investigate how these rates may be influenced by data noise or imperfections in the input data, which are often present in real-world data.

Another outcome worth highlighting involves exact learning, which guarantees to return the highest scoring graph from those explored. The results show that, in practice, exact learning has performed worse than many approximate learning algorithms, and this outcome includes recovering lower scored structures than those recovered by most of the approximate learning algorithms (refer to Figure 6). These results are consistent with those reported in Constantinou et al. (2021), and we propose two possible explanations. Firstly, the often-necessary restriction on the maximum node in-degree hyperparameter of exact learning (it was set to 3 in this paper; i.e., default setting) inevitably limits the ability of exact learning in exploring denser graphs that could have a higher objective score. Secondly, exact learning typically relies on pruning strategies that prune off parts of the search space not containing the highest scoring graph. This guarantee, however, assumes *clean* input data. In practice, real data violate this guarantee. For example, in Constantinou et al. (2021) it was shown that simple learners are more resilient to data noise compared to exact or more sophisticated approximate learners. While exact learning represents an important theoretical exercise, it remains an open question whether exact solutions are useful in practice. Future research could explore whether exact solutions are more accurate than approximate solutions in real settings where the input data contain noise or imperfections.

The overall results also highlight some interesting patterns involving continuous distributions. Specifically, the algorithms that support learning from continuous data are found to learn considerably denser graphs compared to the corresponding graphs they would learn from discretised data. These denser graphs were found to be further away from the knowledge-based causal graph. This result extends to continuous optimisation which was initially viewed positively as a new class of learning through the NOTEARS algorithm (Zheng et al., 2018), but has been shown to be rather unsatisfactory in practice (Constantinou et al., 2021; Kaiser and Sipos, 2022). On this basis, it is likely that structure learning from continuous data not only is subject to a high risk of model overfitting, but also that the many edges recovered will likely be associational rather than causal. A possible future research direction in structure learning from continuous data and continuous optimisation would be to investigate the ratio of spurious edges versus the ratio of true edges that these algorithms recover, especially when the ground truth graph is sparse. It is important to understand why learning from continuous data tends to produce considerably denser graphs compared to the graphs learnt from corresponding discretised data.

Lastly, we note that the traditional causal structure learning algorithms investigated in this paper are oblivious to distribution shifts that may occur over time. This is particularly important when working with time-series data such as COVID-19, where some of the effects captured are known to subject to a time lag. For example, Dehning et al. (2020) show that the effectiveness of COVID-19 policy interventions can only be revealed through change-point detection (i.e., identifying the point in time when one model changes to a new model), and this would explain the inability of most structure learning algorithms tested in this paper to establish a relationship between lockdown and COVID-19 hospitalisations. Few studies have investigated how causal structure learning could account for time-varying distribution shifts in the data (Kummerfeld and Danks, 2013; Kocacoban and Cussens, 2019; Huang et al., 2020; Bregoli et al. (2021)), although none of these algorithms has been adopted by the structure learning community. Glymour et al. (2019) acknowledge that the "*general problem of estimating the causal generating processes from time series is not close to solved*". The problem of recovering time-varying causal structure remains under-investigated. Relevant studies in other areas, and likely those that focus on change-point detection (Fearnhead and Liu, 2007; Adams and MacKay, 2007; Saatçi et al., 2010; Knoblauch et al., 2018) could provide valuable insights for future extensions of existing solutions to this problem.

We conclude that accurate causal structure learning remains an open problem, and that no study should rely on a single structure learning algorithm in determining causal relationships. We also note that the inconsistency in recovering causal structure is not restricted to causal machine learning algorithms, but rather extends to causal knowledge; i.e., each author devised a different knowledge graph from the set of input data variables, before arriving at a consensus. These results highlight the difficulty in determining causal structure, the need to consider multiple sources of information, and the benefits of a model-averaging procedure.

**Credit author statement**

| | Conceptualization | Methodology | Data Curation | Coding | Experiments | Analysis of results | Visualization | Software and repository | Writing - Original Draft | Writing - Review & Editing | Supervision |
|---|---|---|---|---|---|---|---|---|---|---|---|
| Constantinou: | X | X | X | X | X | X | X | X | X | X | X |
| Kitson: | X | X | | X | X | | | | | X | |
| Liu: | X | X | | X | X | | | | | X | |
| Kiattikun: | X | X | | X | X | | | | | X | |
| Hashemzadeh: | X | | | X | X | | | | | X | |
| Nanavati: | X | | | X | X | | | | | X | |
| Mbuvha: | X | | | | | | | | | X | |
| Petrungaro: | X | | | | | | | | | X | |

**Appendix A:** Supplementary information on data collation

**Table A.4.** Manual construction of the *Schools* index that forms part of the *Work and school activity* variable no. 12, and variables 14, 16 and 17 as described in Table 1.

| No. | Name | Details |
|---|---|---|
| n/a | Schools | Introduced states {Open, Partially Open, Closed} consistent with the UK operational guidance for schools during the COVID-19 pandemic (GOV.UK, 2022h; 2022i; 2022j; Wikipedia, 2022b). |
| 14 | Face masks | Introduced states {No, Optional, Yes} consistent with the mask mandates in the UK during the COVID-19 pandemic (GOV.UK, 2022g; Wikipedia, 2022a). |
| 16 | Season | Introduced states {Winter, Spring, Summer, Autumn} consistent with the four seasons. |
| 17 | Majority COVID variant | Introduced states {Initial, Alpha, Delta, Omicron, Omicron BA.2} consistent with the UK Government data on variants (GOV.UK, 2022k) and Our World in Data (Our World in Data, 2022b). |

**Table A.5.** Manual construction of the *Lockdown* (no. 15) variable based on information on lockdown mandates in the UK during the COVID-19 pandemic, provided by the Institute for Government (2022).

| | Government measures | | | | | |
|---|---|---|---|---|---|---|
| Variable state | Full-scale nationwide lockdown | Allowed to attend workplace or school | Non-essential places open | Restrictions on social gatherings | Social distancing restrictions | Masks optional with no mandates |
| No_or_limited_measures | | | | | | ✓ |
| Social_distancing | | | | | ✓ | |
| Weak_lockdown | | | ✓ | ✓ | | |
| Moderate_lockdown | | ✓ | | | | |
| Severe_lockdown | ✓ | | | | | |

**Appendix B:** Average graphs

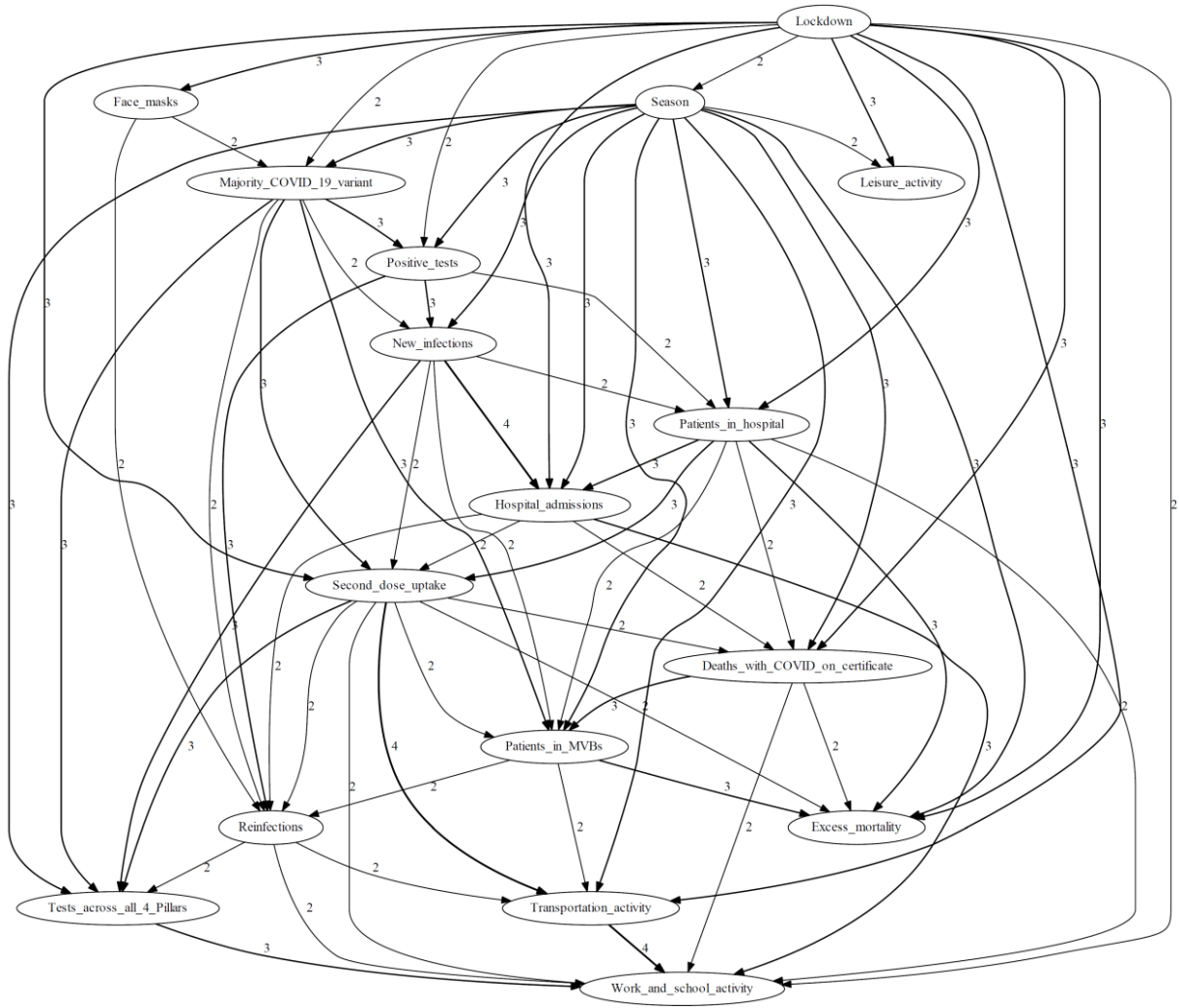

**Figure B.1.** The *All_mixed* average graph obtained across 5 experiments where the input data set contained mixed data. The graph contains a total of 67 edges, where the edge labels represent the number of times the given edge appeared in the 5 outputs considered, and the width of the edges increases with this number. Edges that appeared less than 2 times across the 5 input graphs are not included.

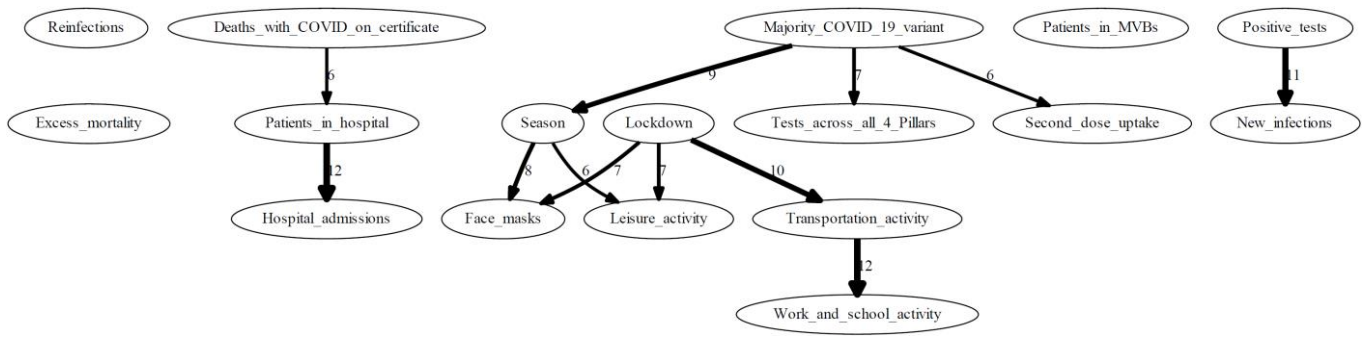

**Figure B.2.** The *All_hybrid* average graph obtained across 17 hybrid learning experiments. The graph contains a total of 12 edges, where the edge labels represent the number of times the given edge appeared in the 17 outputs considered, and the width of the edges increases with this number. Edges that appeared less than 6 times across the 17 input graphs are not included.

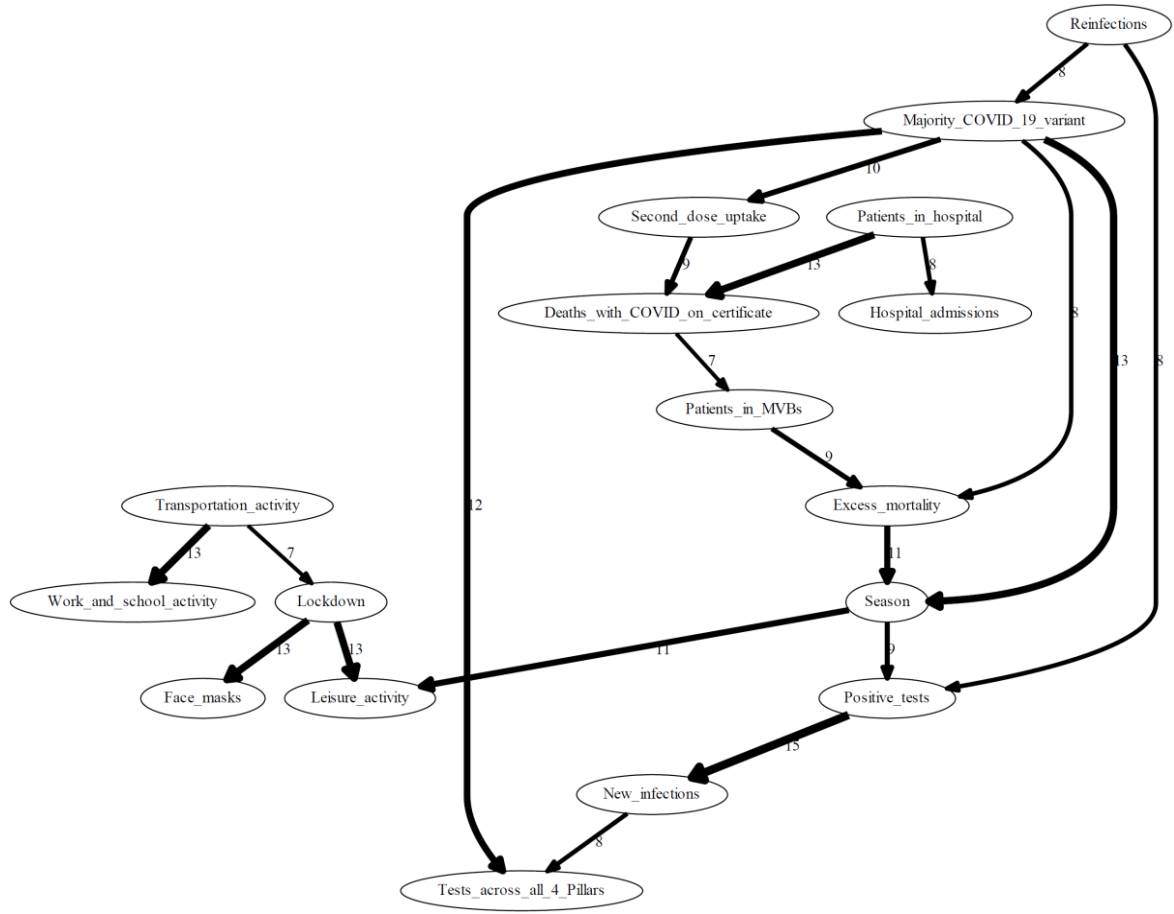

**Figure B.3.** The *All_quartiles* average graph obtained across 19 experiments where the input data set contained mixed data. The graph contains a total of 20 edges, where the edge labels represent the number of times the given edge appeared in the 19 outputs considered, and the width of the edges increases with this number. Edges that appeared less than 7 times across the 19 input graphs are not included.

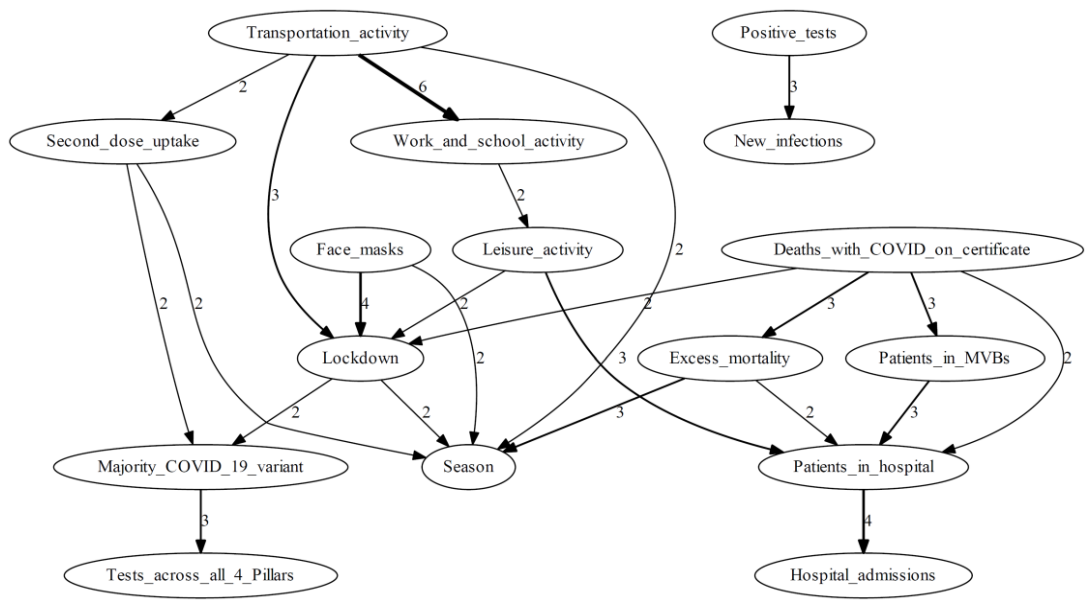

**Figure B.4.** The *All_constraint-based* average graph obtained across 4 constraint-based experiments. The graph contains a total of 23 edges, where the edge labels represent the number of times the given edge appeared in the 4 outputs considered, and the width of the edges increases with this number. Edges that appeared less than 2 times across the 4 input graphs are not included.

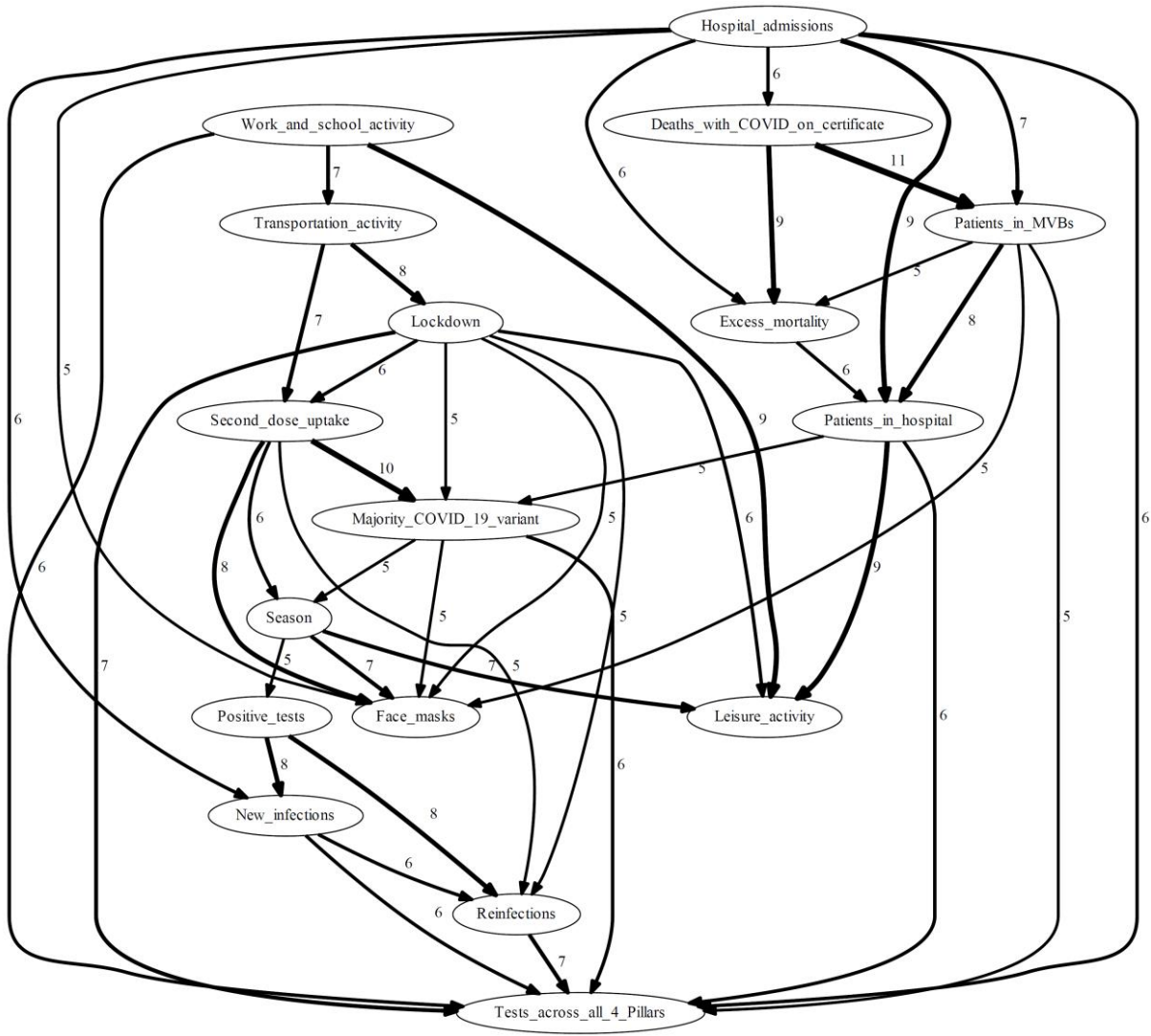

**Figure B.5.** The *All_continuous* average graph obtained across 14 experiments where the input data set contained continuous data. The graph contains a total of 43 edges, where the edge labels represent the number of times the given edge appeared in the 14 outputs considered, and the width of the edges increases with this number. Edges that appeared less than 5 times across the 14 input graphs are not included.

arXiv pre-print, 2023.

arXiv pre-print, 2023.